\documentclass[10pt,twocolumn,letterpaper]{article}

\usepackage[pagenumbers]{wacv} %

\definecolor{wacvblue}{rgb}{0.21,0.49,0.74}
\usepackage{hyperref}
\hypersetup{
  breaklinks=true,
  colorlinks=true,
  allcolors=wacvblue,
  pdftitle={ParticleSplat: Self-supervised Object-centric Latent Particle Splatting},
  pdfauthor={Lyuxing He, Daniel Guo, Elizabeth Terveen, Deepak Pathak, David Held, Tal Daniel}
}

\usepackage{graphicx}
\usepackage{booktabs}
\usepackage{mathrsfs}
\usepackage{pifont}
\usepackage{multirow}
\usepackage{wrapfig}
\usepackage{subcaption}
\usepackage{tabularx}
\usepackage{listings}
\lstdefinestyle{mypython}{
  language=Python,
  basicstyle=\ttfamily\scriptsize,
  keywordstyle=\color{blue},
  commentstyle=\color{teal!70},
  stringstyle=\color{orange},
  showstringspaces=false,
  frame=single,
  rulecolor=\color{black!30},
  numbers=left,
  numberstyle=\tiny\color{gray},
  xleftmargin=2.5em,
  framexleftmargin=2.5em
}

\title{ParticleSplat: Self-supervised Object-centric Latent Particle Splatting}

\author{
Lyuxing He\textsuperscript{1\thinspace\ding{41}},
Daniel Guo\textsuperscript{1},
Elizabeth Terveen\textsuperscript{1},
Deepak Pathak\textsuperscript{1},
David Held\textsuperscript{1},
Tal Daniel\textsuperscript{1}\\[10pt]
\textsuperscript{1}Robotics Institute, Carnegie Mellon University
}

\begin{document}
\maketitle

\begingroup
\renewcommand{\thefootnote}{\ding{41}}
\footnotetext[1]{\raggedright\scriptsize
Corresponding author: Lyuxing He
(\texttt{lyuxingh@andrew.cmu.edu}).}
\endgroup

\begin{abstract}
We present ParticleSplat, a self-supervised object-centric representation learning method that decomposes scenes into a set of latent ``particles'' representing semantic entities through feedforward 3D Gaussian Splatting. Building on the Deep Latent Particles (DLP) framework, which represents images as a set of particles with attributes such as position, scale, and visual appearance, we address a key limitation of DLP: its inherently 2D nature, which prevents explicit
3D spatial and geometric reasoning that are critical for downstream tasks such as robotic manipulation. Leveraging the structural similarity between latent particles and 3D Gaussian primitives, we introduce a 3D latent particle space trained with a novel view synthesis objective. Our model jointly encodes multiple views with camera poses into a shared 3D object-centric latent space, then transforms particles into particle-aligned 3D Gaussians whose composition reconstructs the full scene. On simulated and real-world datasets, we show that this formulation inherently learns object masks without supervision and supports controllable 3D scene editing, such as moving objects by modifying particles in the latent space. We further establish that the learned 3D representation improves downstream performance on robotic manipulation tasks. Videos and code are available: \url{https://lyuxinghe.github.io/ParticleSplat-website/}
\end{abstract}

\section{Introduction}
\label{sec:intro}
3D Gaussian Splatting (3DGS~\cite{kerbl20233d}) has driven much of the recent progress in 3D reconstruction, offering real-time, high-fidelity novel view synthesis from multi-view RGB images via differentiable rendering. While the original formulation optimizes a separate Gaussian representation for each scene, often resulting in millions of primitives and limiting its suitability for downstream applications, recent work has extended 3DGS to neural feedforward models that perform generalizable 3D reconstruction from sets of input images~\cite{ziwen2025longlrm, jiang2025anysplat}. For decision making that could benefit from rich 3D information (e.g., occlusions or fine-grained manipulation), however, typical feedforward approaches do not impose a representation bottleneck and remain high-dimensional, so the potential advantages of 3DGS representations for downstream policy learning in robotic manipulation remain largely underexplored.

In robotic domains, complex multi-object manipulation from visual observations demands scene representations that capture the objects, gripper, and obstacles in a form suitable for policy learning; however, such structured representations are non-trivial to obtain. While recent object-centric approaches have shown promising performance, many rely on supervised annotations~\cite{zhu2023viola, shi2024pocr} from large pre-trained models, which limits their applicability in new or visually diverse environments. As a result, methods that employ \textit{self-supervised} object-centric learning~\cite{haramati2024ecrl, qi2025ecdiffuser} have emerged, demonstrating significant performance gains on long-horizon, multi-object manipulation tasks and exhibiting strong compositional generalization. Yet these representations are typically learned from 2D RGB observations and do not explicitly exploit available 3D information induced from multi-view observations and camera geometry, potentially missing important geometric cues, such as depth, that are critical for robust decision-making in 3D.

Consequently, a complementary line of work directly exploits 3D information, such as point clouds or voxels, showing clear advantages over purely 2D image inputs in settings with severe occlusions and tasks that critically depend on accurate object geometry~\cite{goyal2023rvt, goyal2024rvt, ze20243d, wang2024equivariant, shridhar2023perceiver, gervet2023act3d, zhang20263ddlp}. In parallel, several recent methods propose to use Gaussian primitives learned via a 3DGS objective~\cite{ze2023gnfactor, lu2024manigaussian, zhang2026gem3d} as the underlying scene representation for downstream manipulation from multi-view RGB or RGBD observations, demonstrating that 3D Gaussians can serve as powerful, task-relevant features. However, both of these directions typically operate directly in high-dimensional 3D spaces, which necessitates memory-efficient architectures, increases training and inference costs, and does not exploit structured object-centric bottlenecks that could improve data efficiency, generalization, and interpretability.

In this work, we aim to bridge this gap by learning self-supervised object-centric representations that explicitly embed 3D information in their latent space through a 3D Gaussian splatting objective. Interestingly, the recent self-supervised object-centric framework Deep Latent Particles (DLP~\cite{daniel2024ddlp}), which represents scenes as sets of latent particles, exhibits a close structural alignment with the Gaussian primitives used in 3DGS: both associate each element with geometric coordinates (position) and transparency-like attributes, but DLP particles are inherently 2D whereas Gaussian primitives are defined in 3D. Motivated by this connection, we introduce \textbf{ParticleSplat}, the first self-supervised object-centric feedforward Gaussian splatting model. Our approach builds on DLP by extending the particle encoder to jointly process multi-view inputs and camera poses, and by replacing the pixel-based decoder with a differentiable renderer applied to 3D Gaussians obtained via a learned transformation from latent particles, and aligned to them, thereby propagating 3D structure back into the latent space. We show that this yields a controllable, compact, object-centric 3D latent representation that supports scene editing (e.g., moving or reconfiguring objects in 3D) and enables more efficient and higher-performing downstream robotic manipulation policies.

Our contributions are as follows: (1) we propose ParticleSplat, the first---to the best of our knowledge---end-to-end self-supervised object-centric latent representation trained via feedforward 3D Gaussian splatting; (2) we demonstrate higher-fidelity novel view synthesis on both simulated and real-world datasets compared to similar representation-oriented baselines; (3) we show that the disentangled, compositional latent space enables intuitive scene editing, such as moving objects in 3D and rendering modified novel views; and (4) we demonstrate that our learned 3D object-centric representation improves performance on complex downstream robotic manipulation tasks compared to 2D object-centric representations and non-object-centric Gaussian splatting models.

\section{Related Work}
\label{sec:rel_work}
We provide an overview of the three main research directions related to our work: self-supervised object-centric representations, feedforward 3D Gaussian splatting for efficient view synthesis, and scene reconstruction methods designed to support downstream decision-making.
\textbf{Self-supervised Object-centric Representations}: Recent self-supervised object-centric methods for RGB-based image and video decomposition can be broadly categorized into three families: patch-based~\cite{lin2020space, crawford2019spair}, slot-based~\cite{burgess2019monet, greff2019iodine, locatello2020slotattn}, and particle-based~\cite{daniel22adlp, daniel2024ddlp, daniel2026lpwm}. These approaches typically encode a set of latent entities from the input image and reconstruct it by composing their decoded representations. Our method extends the particle-based DLP framework~\cite{daniel2026lpwm} to encode multi-view RGB or RGBD observations into 3D entities, which are subsequently transformed into 3D Gaussians that render pixel-space novel views, instead of relying on a decoder that directly predicts pixels. Several recent methods have also extended self-supervised object-centric representations to novel view synthesis, where new views are generated directly by a pixel-space decoder, using either patch-based~\cite{chen2021roots} or slot-based~\cite{li2020mulmon, yuan2022ocloc, sajjadi2022osrt, jabri2024dorsal} decompositions. In contrast, our approach integrates feedforward 3D Gaussian Splatting into the DLP framework, such that novel views are rendered from an explicitly 3D latent space. Other works have explored neural radiance fields (NeRFs) as the underlying scene representation~\cite{smith2023colf, yu2022uorf, stelzner2021obsurf, luo2025uocf}, but these typically incur higher inference-time costs because they require querying scene content at many continuous coordinates, whereas splatting-based methods enable more efficient, rasterization-based rendering.

\textbf{Feedforward 3D Gaussian Splatting}:
Instead of optimizing Gaussians per scene~\cite{kerbl20233d}, feedforward Gaussian Splatting methods learn predictors that directly regress Gaussian primitives from a set of context views. Early works predict pixel-aligned Gaussians from posed multi-view inputs by leveraging epipolar attention~\cite{charatan2024pixelsplat} or plane-sweeping cost volumes~\cite{chen2024mvsplat}. DepthSplat~\cite{xu2025depthsplat} fuses monocular features into cost-volume formulations to explicitly regress depth maps for Gaussian initialization, while VolSplat~\cite{wang2025volsplat} uses geometry-aware features to populate a sparse 3D voxel grid. Although such architectures achieve impressive visual fidelity and scalability, they bake explicit cross-view correlation inductive biases (e.g., cost volumes, epipolar constraints) into the model and are primarily designed as general-purpose reconstruction systems rather than as compact, interpretable representations for control. LatentSplat~\cite{wewer2024latentsplat} regresses a set of 3D Gaussians from posed input views with deterministic geometry and spherical harmonic (SH) RGB coefficients, placing a variational distribution only over latent SH feature coefficients. The variational modeling is largely confined to appearance, and the resulting latent codes are not demonstrated to support downstream decision-making or object-centric control. In contrast, our goal is not to compete with large-scale feedforward 3DGS methods on raw visual fidelity, but to learn a compact, structured, and controllable latent representation that is directly useful for decision making. To this end, our method maximizes a modified evidence lower bound (ELBO) tailored to a fully variational particle-based representation that avoids explicit geometric constructs, yielding an object-centric latent space that is interpretable and amenable to downstream robotic manipulation policies.

\textbf{Scene Reconstruction for Decision-making}:
Visual imitation learning methods typically mimic expert demonstrations by encoding observations into latent feature vectors; however, a growing subset of approaches enhances policy learning by explicitly training these representations through scene reconstruction. Neural radiance fields (NeRFs)~\cite{mildenhall2021nerf} have been explored in diverse policy backbones ranging from reinforcement learning algorithms~\cite{driess2022reinforcement, shim2023snerl} to transformer-based behavior cloning agents~\cite{ze2023gnfactor}. Extending this direction, KP-NeRF~\cite{wang2022dynamical} moved beyond static geometry by conditioning the radiance field on keypoints to model scene dynamics, enabling prediction of future states for model-based control. Despite these advances, NeRFs often suffer from high rendering latency; consequently, 3D Gaussian Splatting (3DGS,~\cite{kerbl20233d}) has emerged as an alternative, offering real-time rendering speeds while maintaining explicit geometric structure. Recent work integrating 3DGS in decision-making includes language-guided grasping~\cite{zheng2024gaussiangrasper, ji2024graspsplats} and dynamic prediction~\cite{lu2024manigaussian, liu2025spatial, lu2025gwm, zhang2026gem3d}. In this work, we leverage the expressivity of 3DGS to learn a structured object-centric representation latent space, and demonstrate its efficacy in manipulation tasks that require compositional reasoning.

\section{Background}
\label{sec:bg}
In this work, we introduce 3D rendering to particle-based self-supervised object-centric representations. Below, we provide background on the main foundations of our method: Deep Latent Particles (DLP) and 3D Gaussian Splatting (3DGS).

\textbf{Deep Latent Particles (DLP~\cite{daniel2026lpwm})}\footnote{DLP was originally introduced in \cite{daniel22adlp}, then improved to DLPv2~\cite{daniel2024ddlp}, and recently updated to DLPv3~\cite{daniel2026lpwm}. We use the latter and simply refer to it as DLP.}: a variational autoencoder (VAE~\cite{kingma2014autoencoding}) based self-supervised object-centric model that provides a structured, disentangled latent space. For a given image $I \in \mathbb{R}^{C \times H \times W}$, where $C$, $H$, and $W$ denote the number of channels, height, and width respectively, DLP encodes it into a set of $L$ foreground latent particles and a single background particle. Each foreground particle is defined as
\[
z_{\text{fg}} = [z_p, z_s, z_d, z_t, z_f] \in \mathbb{R}^{6 + d_{\text{obj}}},
\]
where each component encodes a disentangled stochastic attribute: position $z_p \sim \mathcal{N}(\mu_p, \sigma_p^2) \in \mathbb{R}^2$, representing the 2D keypoint coordinates; scale $z_s \sim \mathcal{N}(\mu_s, \sigma_s^2) \in \mathbb{R}^2$, representing bounding-box dimensions; depth $z_d \sim \mathcal{N}(\mu_d, \sigma_d^2) \in \mathbb{R}$, specifying compositing order to indicate occlusion relationships; transparency $z_t \sim \mathrm{Beta}(a, b) \in [0,1]$, controlling per-particle opacity; and visual features $z_f \sim \mathcal{N}(\mu_f, \sigma_f^2) \in \mathbb{R}^{d_{\text{obj}}}$, encoding appearance of the local region around the particle (summing to $6 + d_{\text{obj}}$ dimensions). The background is represented by a single particle $z_{\text{bg}} \sim \mathcal{N}(\mu_{\text{bg}}, \sigma_{\text{bg}}^2) \in \mathbb{R}^{d_{\text{bg}}}$,
fixed at the image center and modeling background visual features. DLP learns this decomposition through a compositing operation where each particle is decoded into a spatial appearance map, which is then composited based on position, scale, depth, and transparency to reconstruct the input image.

\textbf{3D Gaussian Splatting (3DGS~\cite{kerbl20233d})}: represents a 3D scene as a set of $N$ anisotropic 3D Gaussians $\mathcal{G}=\{(\boldsymbol\mu_i,\Sigma_i,\mathbf{c}_i,\alpha_i)\}_{i=1}^N$, where $\mu_i\in\mathbb{R}^3$ is the mean position,
$\Sigma_i\in\mathbb{R}^{3\times 3}$ is the covariance, $\mathbf{c}_i$ denotes appearance parameters (e.g., RGB or low-order spherical-harmonic coefficients), and $\alpha_i\in(0,1)$ is opacity. To ensure $\Sigma_i$ remains positive semi-definite, it is factorized into a diagonal scaling matrix $\mathbf{S_i} \in \mathbb{R}^{3 \times 3}$ and a rotation matrix $\mathbf{R_i} \in \mathbb{R}^{3 \times 3}$, such that $\Sigma_i = \mathbf{R_i} \mathbf{S_i} \mathbf{S_i}^T \mathbf{R_i}^T$. To render the scene from a given camera view $\mathbf{W}$, each 3D Gaussian primitive is projected onto the image plane. The 3D mean $\boldsymbol{\mu}_i$ is projected to 2D screen coordinates $\boldsymbol{\mu}'_i$, while the covariance is transformed to a 2D planar covariance $\mathbf{\Sigma}'_i = \mathbf{J} \mathbf{W} \mathbf{\Sigma}_i \mathbf{W}^T \mathbf{J}^T$ using the viewing transform and the affine Jacobian approximation $\mathbf{J}$.
The screen-space influence of the $i$-th Gaussian at pixel $\mathbf{x}$ is then modeled as $G'_i(\mathbf{x}) = \exp \left( -\frac{1}{2} (\mathbf{x} - \boldsymbol{\mu}'_i)^T \mathbf{\Sigma}'^{-1}_i (\mathbf{x} - \boldsymbol{\mu}'_i) \right)$. The final pixel color $\mathbf{C}(\mathbf{x})$ and depth $D(\mathbf{x})$ are computed using differentiable front-to-back alpha compositing. Letting $\alpha_i(\mathbf{x}) = \sigma(o_i) G'_i(\mathbf{x})$ be the effective opacity and $T_i(\mathbf{x}) = \prod_{j<i} (1 - \alpha_j(\mathbf{x}))$ be the transmittance, the rendering is formulated as:
\[
    \mathbf{C}(\mathbf{x}) = \sum_{i \in \mathcal{N}} T_i(\mathbf{x}) \alpha_i(\mathbf{x}) \mathbf{c}_i, \qquad
    D(\mathbf{x}) = \sum_{i \in \mathcal{N}} T_i(\mathbf{x}) \alpha_i(\mathbf{x}) z_i,
\]
where $z_i$ is the depth of the Gaussian center in camera space. In this work, rather than adhering to the standard paradigm of optimizing $\mathcal{G}$ via iterative per-scene gradient descent, we employ a generalizable feedforward network~\cite{charatan2024pixelsplat, chen2024mvsplat} to directly regress Gaussian primitives from 2D observations.

\section{ParticleSplat: Self-supervised Object-centric Latent Particle Splatting}
\label{sec:method}
We seek to learn a structured 3D latent representation from a set of $V$ posed 2D observations $\mathcal{O} = \{ (\mathbf{x}^{v}, \mathbf{P}^{v}, \mathbf{K}^{v}) \}_{v=1}^{V}$. For each view $v$, the camera extrinsic pose is $\mathbf{P}^v = [\mathbf{R}^v \mid \mathbf{t}^v] \in SE(3)$, with rotation $\mathbf{R}^v \in SO(3)$ and translation $\mathbf{t}^v \in \mathbb{R}^3$, and the camera intrinsics are given by $\mathbf{K}^v \in \mathbb{R}^{3 \times 3}$. Each observation $\mathbf{x}^{v} \in \mathbb{R}^{C \times H \times W}$ has dimensions $H, W$ and $C \in \{3, 4\}$ channels for RGB or RGB-D inputs, respectively. This formulation supports both multi-view and monocular setups, with monocular RGB-D ($V=1$, $C=4$) strictly required to resolve scale ambiguity where object size and depth are indistinguishable without parallax.

\begin{figure*}[!ht]
\vspace{-2em}
\vskip 0.2in
\begin{center}
\centerline{\includegraphics[width=0.8\textwidth]{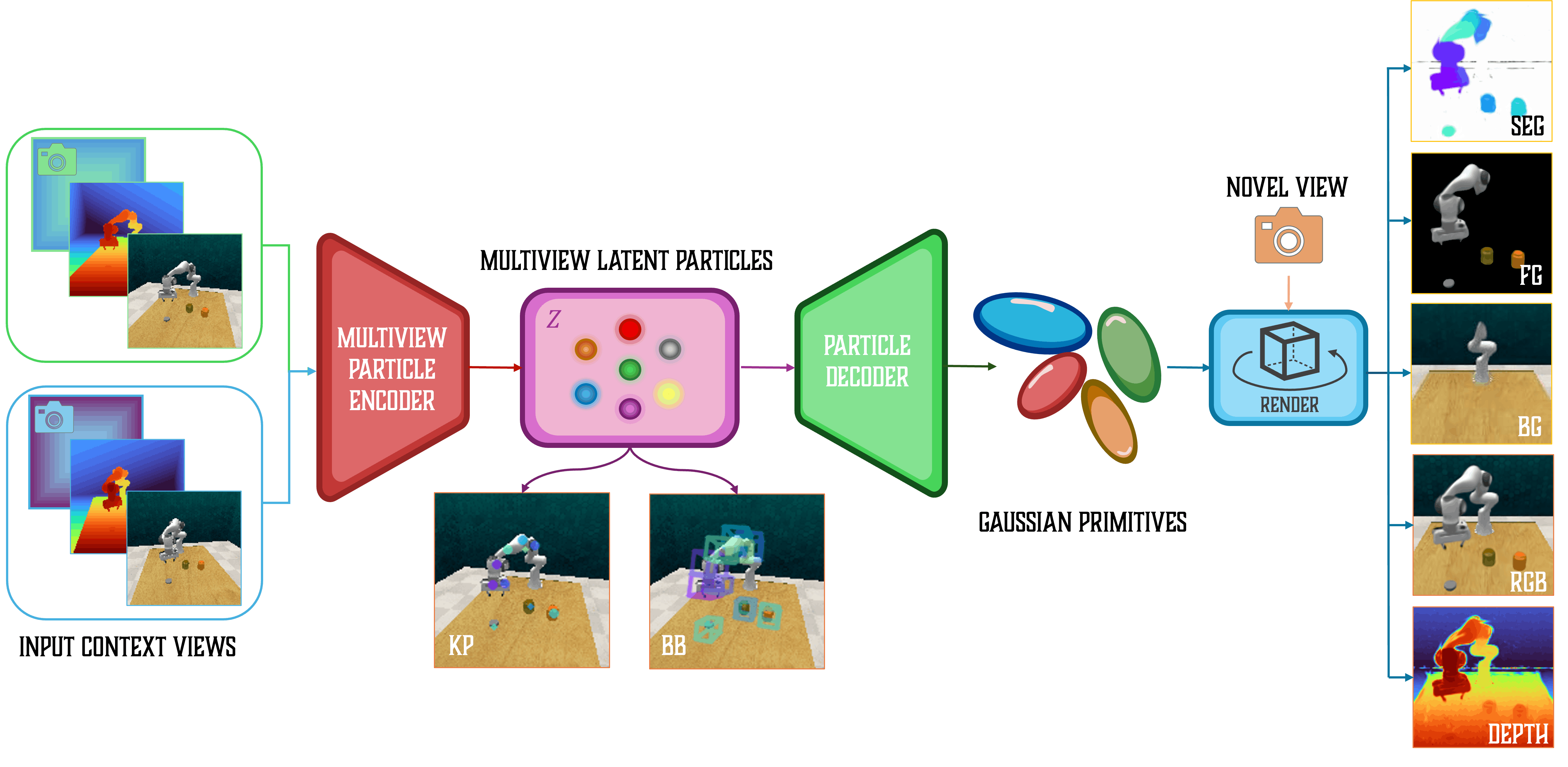}}
\vspace{-0.5em}
\caption{
\textbf{ParticleSplat architecture.} Given multi-view context observations (RGB images, camera poses encoded as Plücker ray coordinates, and optional depth maps), the view-aware encoder infers a set of object-centric latent particles. The decoder transforms these particles into world-space 3D Gaussian primitives, enabling compositional novel view synthesis. The object-centric structure naturally provides per-particle segmentation masks (SEG) alongside learned geometric attributes such as keypoints (KP) and bounding boxes (BB).}
\label{fig:arch}
\end{center}
\vskip -0.2in
\vspace{-1.5em}
\end{figure*}

Crucially, the particle structure in Deep Latent Particles (DLP)—with explicit position, scale, depth, transparency, and appearance attributes—closely mirrors the attributes of 3D Gaussian primitives in 3DGS, but remains confined to 2D image space. We bridge this gap by extending DLP particles to 3D scene space, enabling self-supervised learning of geometry-aware, object-centric representations that naturally yield segmentation masks without explicit supervision. Our formulation adopts a self-supervised VAE with particle-based latents: we define a multi-view particle representation, design a view-aware encoder that fuses 2D observations with camera geometry to infer particle posteriors, and introduce a Gaussian-splatting decoder that transforms particles into 3D Gaussians and renders them via differentiable rasterization, optimized under a view-aware geometric ELBO. In the main text, we focus on this decoder and training objective as the core contribution, and refer the reader to Appendices~\ref{subsec:apndx_method} for extended details on the multi-view latent particles, encoder, decoder, and loss. Together, this yields the first self-supervised object-centric VAE for 3D Gaussian Splatting, producing compact particle representations that enable both high-fidelity novel view synthesis and interpretable control for downstream decision-making. Our model is illustrated in Figure~\ref{fig:arch}.

\textbf{Multi-view Latent Particles (Appendix~\ref{sec:method_particles}).} We structure the latent space as a set of multi-view particles
\(
\mathcal{Z} =
\{z_{\mathrm{fg}}^{v,\ell}\}_{v=1,\ell=1}^{V,L}
\;\cup\;
\{z_{\mathrm{bg}}^{v}\}_{v=1}^{V},
\)
consisting of \(V \times (L + 1)\) entities, where \(V\) is the number of views, \(L\) is the number of foreground particles, and each view is augmented with a single background particle. Each foreground particle
\(z_{\text{fg}} = [z_p, z_s, z_d, z_e, z_t, z_f] \in \mathbb{R}^{7 + d_{\text{obj}}}\)
encodes disentangled stochastic attributes that explicitly separate \emph{image-plane} geometry (projected 2D appearance in the camera view) from \emph{volumetric} structure (3D extent in world coordinates). The attributes follow DLP~\cite{daniel2026lpwm} as described in Sec.~\ref{sec:bg},
with one key 3D extension: in addition to depth $z_d \sim \mathcal{N}(\mu_d, \sigma_d^2) \in \mathbb{R}$ (encodes distance from camera along the viewing ray and determines compositing order), we introduce \emph{depth extent}
$z_e \sim \mathcal{N}(\mu_e, \sigma_e^2) \in \mathbb{R}$,
which controls the particle's thickness along the viewing direction---e.g., a large $z_e$ creates ``deep'' objects spanning multiple depth planes (like a box), while a small $z_e$ yields ``flat'' objects concentrated near $z_d$ (like a sheet).

\textbf{Encoder (Appendix~\ref{sec:method_encoder}).} The encoder $q_{\phi}(\mathcal{Z} \mid \mathcal{O})$ models the posterior distribution over multi-view latent particles given posed observations. We build on the hierarchical particle encoder of DLP~\cite{daniel2026lpwm}, which extracts keypoint proposals via differentiable spatial softmax~\cite{jakab2018unsupervised}, crops local glimpses with a Spatial Transformer Network (STN~\cite{jaderberg2015stn}), and predicts particle attributes using small CNNs. An attention-based interaction module then refines depth and appearance through cross-view attention over the particles of all context views.
Similarly to recent feedforward Gaussian splatting models~\cite{zhang2024gslrm,ziwen2025longlrm,jin2025lvsm,charatan2024pixelsplat,chen2024mvsplat},  we condition particle features on camera geometry to relate observations across views. For each pixel (and hence each particle location), we compute its associated camera ray from intrinsics and extrinsics, and encode it using Plücker coordinates $\mathbf{r} \in \mathbb{R}^6$~\cite{zhang2024cameras}, which provide a continuous, translation-invariant representation of 3D lines. These ray embeddings are injected into convolutional feature maps via spatially aligned projection networks and into particle embeddings via FiLM modulation~\cite{perez2018film} before self-attention, enabling the encoder to infer consistent 3D particle distributions from 2D views. Furthermore, extending DLP to the single-view RGB+D setting (to resolve scale ambiguity) requires concatenating depth as a 4th input channel and adapting the CNNs accordingly. While effective, we found that processing RGB and depth features separately via parallel CNNs then fusing multi-level features yields superior performance. Full architectural details are provided in Appendix~\ref{sec:method_encoder}.

\textbf{Decoder (Appendix~\ref{sec:method_decoder}).} The decoder $p_{\theta}(\mathbf{x} \mid \mathcal{Z})$ replaces the pixel-space decoder of DLP with a 3D Gaussian renderer: it maps latent particles to a set of world-space 3D Gaussian primitives and renders them with a differentiable 3DGS rasterizer~\cite{kerbl20233d}. Given $V$ posed cameras $\{(P^v, K^v)\}_{v=1}^{V}$ and the multi-view latent particle set
$\mathcal{Z}
=
\{z_{\mathrm{fg}}^{v,\ell}\}_{v=1,\ell=1}^{V,L}
\;\cup\;
\{z_{\mathrm{bg}}^{v}\}_{v=1}^{V}$,
the decoder produces per-view RGB and depth reconstructions $\{(\hat{\mathbf{I}}^{v},\hat{\mathbf{D}}^{v})\}_{v=1}^{V}$ by decoding each particle into a local Gaussian field, lifting all Gaussians to world space, and then “stitching’’ them into a single scene representation for rendering. In practice, both the foreground and background decoders are implemented as CNNs with dedicated Gaussian heads.

\textbf{Foreground particles.} The foreground decoder $D_{\mathrm{fg}}$ first maps the visual features $z_f^{v,\ell}$ to a canonical patch of $N=S_h S_w$ Gaussians in a local 2D coordinate frame:
\[
\mathscr{G}^{v,\ell}_{\mathrm{fg}}
=
D_{\mathrm{fg}}(z_{f}^{v,\ell})
=
\{\mathbf{c}_i^{v,\ell},\boldsymbol{\delta}_i^{v,\ell},\Delta d_i^{v,\ell},\mathbf{q}_i^{v,\ell},\mathbf{s}_i^{v,\ell},o_i^{v,\ell}\}_{i=1}^{N},
\]
where each Gaussian has SH color coefficients $\mathbf{c}_i^{v,\ell}$, a sub-cell offset $\boldsymbol{\delta}_i^{v,\ell}$, a depth residual $\Delta d_i^{v,\ell}$, a quaternion $\mathbf{q}_i^{v,\ell}$, anisotropic scales $\mathbf{s}_i^{v,\ell}$, and an opacity logit $o_i^{v,\ell}$.

We then apply the particle’s affine transform (parameterized by position and scale $(z_p^{v,\ell},z_s^{v,\ell})$) to move these Gaussians from the canonical patch frame into the full image plane. Concretely, canonical grid coordinates and sub-cell offsets are combined to give patch-normalized centers, which are mapped through the affine transform to image-normalized coordinates; the same affine map is applied to Gaussian covariances via its Jacobian, ensuring that ellipsoids are consistently scaled and placed in the image domain. At this point, each foreground Gaussian has a well-defined image-plane center and covariance, while its SH color, depth residual, and opacity logit remain unchanged.

Next, we lift these image-plane Gaussians into world-space 3D Gaussians using the particle’s depth attributes and camera geometry. The depth and extent latents $(z_d^{v,\ell}, z_e^{v,\ell})$ are converted into a bounded planar-$Z$ depth interval, and each Gaussian receives a depth $\zeta_i^{v,\ell}$ within $[d_{\min}, d_{\max}]$ by adding a bounded residual $\Delta d_i^{v,\ell}$. Given the pixel-normalized center $\mathbf{u}_i^{v,\ell}$ and camera pose $(P^v, K^v)$, we compute the camera ray and place the Gaussian mean at
\[
\boldsymbol{\mu}_i^{v,\ell} = \mathbf{o}^v + \zeta_i^{v,\ell}\,\mathbf{d}^v(\mathbf{u}_i^{v,\ell}),
\]
where $\mathbf{o}^v$ is the camera center and $\mathbf{d}^v(\cdot)$ is the unnormalized ray direction. Covariances are formed from bounded scales and normalized quaternions in a camera-aligned frame and rotated into world coordinates; opacities pass $o_i^{v,\ell}$ through a bounded nonlinearity gated by the transparency $z_t^{v,\ell}$, and SH coefficients are rotated into the world frame. This yields a set of world-space 3D Gaussians
$\mathcal{G}^{v,\ell}_{\mathrm{fg}}
=
\{(\boldsymbol{\mu}_i^{v,\ell}, \Sigma_i^{v,\ell}, \mathbf{c}_i^{v,\ell}, \alpha_i^{v,\ell})\}_{i=1}^{N}$
for each foreground particle.

\textbf{Background particle.}
For each view $v$, a background particle $z_{\mathrm{bg}}^{v}$ is decoded by a background decoder $D_{\mathrm{bg}}$ into a dense field of pixel-aligned Gaussians at resolution $H\times W$,
$\mathscr{G}^{v}_{\mathrm{bg}} = D_{\mathrm{bg}}(z_{\mathrm{bg}}^{v})$,
parameterized analogously to the foreground case. Since these Gaussians are already defined in the full image plane, they do not require the patch-to-image affine unwarping; we directly lift them to world-space 3D Gaussians using the same ray- and depth-based mapping as for foreground Gaussians, obtaining $\mathcal{G}^{v}_{\mathrm{bg}}$.

\textbf{Stitching and rendering.}
The foreground and background Gaussians from all views are concatenated into a single scene set $\mathcal{G}$; this simple ``stitching'' suffices thanks to the compositional nature of Gaussian splatting (Appendix~\ref{sec:method_decoder}).
Given any target camera $(P^{t},K^{t})$, we render RGB and depth via the standard 3DGS rasterizer with front-to-back alpha compositing,
$(\hat{\mathbf{I}}^{t}, \hat{\mathbf{D}}^{t}) = \mathcal{R}(\mathcal{G}; P^{t}, K^{t})$,
enabling multi-view reconstruction and novel view synthesis directly from the compact object-centric latent particles. Furthermore, because each foreground particle corresponds to a subset of Gaussians, we can render particles independently to obtain explicit 3D object decompositions and controllable scene edits; we detail these per-particle renderings and editing operations in Appendix~\ref{sec:method_decoder}.

\textbf{Loss (Appendix~\ref{sec:method_elbo}).} Given a set of posed observations $\mathcal{O}=\{(\mathbf{x}^v,\mathbf{P}^v,\mathbf{K}^v)\}$, we partition the views into a context set $\mathcal{C}$ and a target set $\mathcal{T}$. Context views are encoded to multi-view latent particles $\mathcal{Z}$, while target views supervise the Gaussian-splatting decoder and encourage 3D consistency and occlusion reasoning. We optimize a multi-view ELBO that treats $\mathcal{Z}$ as describing a \emph{single} underlying 3D scene and each $\mathbf{x}^v$ as a 2D projection of that scene.

Concretely, we average reconstruction terms over all context and target views, but apply a single scene-level KL term that is not normalized by the number of views:
\begin{equation*}
\begin{aligned}
\mathcal{L}_{\text{ELBO}}
&=
\underbrace{\frac{1}{|\mathcal{C}|+|\mathcal{T}|}
\Big(
\beta_{\mathrm{rec}}^{\mathrm{ctx}}\!\!\sum_{v\in\mathcal{C}}\!\!
\mathcal{E}(\mathbf{x}^v, \hat{\mathbf{x}}^v)
+
\beta_{\mathrm{rec}}^{\mathrm{tgt}}\!\!\sum_{t\in\mathcal{T}}\!\!
\mathcal{E}(\mathbf{x}^t, \hat{\mathbf{x}}^t)
\Big)}_{\mathcal{L}_{\mathrm{rec}}}
\\
&\quad+
\beta_{\mathrm{KL}}\,
\underbrace{\mathrm{KL}\!\left(
q_{\phi}(\mathcal{Z}\mid\mathcal{C})
\,\Vert\,
p(\mathcal{Z})
\right)}_{\mathcal{L}_{\mathrm{KL}}}.
\end{aligned}
\end{equation*}
where $\beta_{\mathrm{rec}}^{\mathrm{ctx}}$ and $\beta_{\mathrm{rec}}^{\mathrm{tgt}}$ weight context and target reconstructions, and $\beta_{\mathrm{KL}}$ controls the strength of scene-level regularization. For each view $i$, the reconstruction compares the ground-truth RGB–depth pair $\mathbf{x}^{i}=(\mathbf{I}^{i},\mathbf{D}^{i})$ with its rendering $\hat{\mathbf{x}}^{i}=(\hat{\mathbf{I}}^{i},\hat{\mathbf{D}}^{i})$:
\[
\mathcal{E}(\mathbf{x}^{i}, \hat{\mathbf{x}}^{i})
=
\mathcal{L}_{\mathrm{img}}^{(i)}
+
\lambda_{D}^{(i)}\,\big\|\tilde{\mathbf{D}}^{i}-\tilde{\hat{\mathbf{D}}}^{i}\big\|_2^2,
\]
where $\lambda_{D}^{(i)}$ balances depth supervision relative to the image loss\footnote{Our model can be trained end-to-end with multi-view RGB observations without explicit depth maps, as the splatting objective and Plücker ray conditioning provide sufficient geometric supervision. Depth is used only optionally for RGB-D inputs.}. The image term $\mathcal{L}_{\mathrm{img}}^{(i)}$ is either a pixel-wise MSE or a VGG-based perceptual loss~\cite{hoshen2019perceptualloss}, depending on the experiment. We provide the full loss details in Appendix~\ref{sec:method_elbo}.

\textbf{Training and Implementation Details.} 
We implement our method in PyTorch~\cite{paszke2017pytorch} using the Adam~\cite{adam_14} optimizer with a batch size of 8. Unless otherwise specified, we use an initial learning rate of $8e^{-5}$ and train until convergence. Training is performed on a single A6000 GPU and takes approximately 30 hours for RLBench dataset, and 2 hours for the real-world dataset. Full training details and hyperparameters are provided in Appendix~\ref{subsec:apndx_hp}. We additionally provide a representation compactness and runtime analysis in Appendix~\ref{subsec:comp-effic}, showing that ParticleSplat maintains a compact particle bottleneck while supporting practical policy learning with low memory usage and real-time latency.

\section{Experiments}
\label{sec:exp}
We conduct experiments with four primary goals: (1) evaluate ParticleSplat's object-centric scene decomposition and novel-view synthesis capabilities against non-object-centric, representation-oriented baselines of comparable scale (Sec.~\ref{subsec:exp_novelv} and~\ref{subsec:exp_ctrl}); (2) demonstrate the interpretability and controllability of the learned representation through latent space modifications and rendering (Sec.~\ref{subsec:exp_ctrl}); (3) ablate the contributions of our novel components within the DLP framework (Sec.~\ref{subsec:exp_novelv}); and (4) demonstrate the efficacy of the learned representation for downstream complex robotic manipulation tasks (Sec.~\ref{subsec:exp_imit}).

\textbf{Datasets.} 
We evaluate on both synthetic and real-world datasets to evaluate novel-view synthesis and assess the applicability of our learned representations to downstream tasks. Our primary synthetic benchmark consists of 10 scenes with a total of 166 variations and 200 episodes of multi-view demonstrations from the RLBench robotic manipulation benchmark~\cite{james2020rlbench} (approximately 660k frames). We also report additional imitation learning results on MimicGen~\cite{mandlekar2023mimicgen} against a broader set of baselines for a holistic comparison. Details on the specific tasks are provided in Appendix~\ref{subsec:apndx_rlbench}. We train a single unified model as the representation shared across all tasks. For real-world experiments, we setup multi-object robotic manipulation scenes and collect multi-view videos with a total amount of 3,000 frames using a mobile phone. We provide a simple, reproducible data preparation recipe in Appendix~\ref{subsec:apndx_recipe}.

\textbf{Baselines.} We compare against methods that use novel view synthesis with Gaussian splatting or NeRF for robotic manipulation representations: ManiGaussian~\cite{lu2024manigaussian} and GNFactor~\cite{ze2023gnfactor}. ManiGaussian learns a 3D voxel representation that predicts spatiotemporal semantic Gaussian primitives, while GNFactor builds a NeRF-style 3D semantic feature field using large pretrained 2D foundation models. Both require multi-view supervision during training but operate from single RGB-D observations at inference, matching our evaluation protocol. Details are provided in Appendix~\ref{subsec:apndx_baselines}.

We also introduce VAE-GS (Appendix~\ref{subsec:vaegs}), an ablation baseline that retains our CNN-based 3D-aware encoder and Gaussian decoder but replaces the object-centric particle latent with a non-structured patchified image latent. VAE-GS uses the same ELBO formulation, differing only in latent parameterization and KL regularization. 

Finally, we include Deep Latent Particles (DLP~\cite{daniel2026lpwm}) as a strong self-supervised object-centric baseline trained on 2D images (Sec.~\ref{sec:bg}). We focus these targeted comparisons on representation learning methods suitable for control rather than large-scale reconstruction methods, which do not produce latents for downstream policy integration.

\subsection{Novel View Synthesis}
\label{subsec:exp_novelv}
We first demonstrate that ParticleSplat captures consistent 3D scene structure through novel view synthesis on synthetic RLBench environments and real-world multi-object scenes. Large novel view synthesis models that impose no representation bottleneck~\cite{charatan2024pixelsplat, chen2024mvsplat, xu2025depthsplat, wang2025volsplat} are orthogonal to our scope. We instead use novel view synthesis as a diagnostic of the representation. Accurate rendering from held-out views indicates that the compact particle bottleneck has captured view-consistent 3D geometry rather than merely fitting image-space appearance, which is what enables decomposition, controllability, and downstream manipulation.

\textbf{Setup.} For RLBench, we use the front camera alongside a NeRF-style trajectory of 21 posed views for supervision and evaluation. In the single-view setting, models receive only the front RGB-D observation; in the multi-view setting, two RGB-D views are randomly sampled from the trajectory as context. Evaluation renders four held-out views of the same trajectory, fixed across all methods for fairness. For real-world scenes, we focus on the multi-view RGB-only setting and evaluate on every fourth view of the capture trajectory. Images are downsampled to $128\times128$, following prior work~\cite{lu2024manigaussian,ze2023gnfactor}, so that a single representation serves both novel view synthesis and imitation learning. RE10K~\cite{zhou2018stereo} results at $256\times256$ are in Appendix~\ref{subsec:nvs-extended}.

\textbf{Metrics.} We report standard novel view synthesis metrics averaged across all evaluation views and scenes: PSNR, SSIM~\cite{wang2004ssim}, and LPIPS~\cite{zhang2018lpips}.

\begin{figure}
    \centering
    \includegraphics[width=1.0\linewidth]{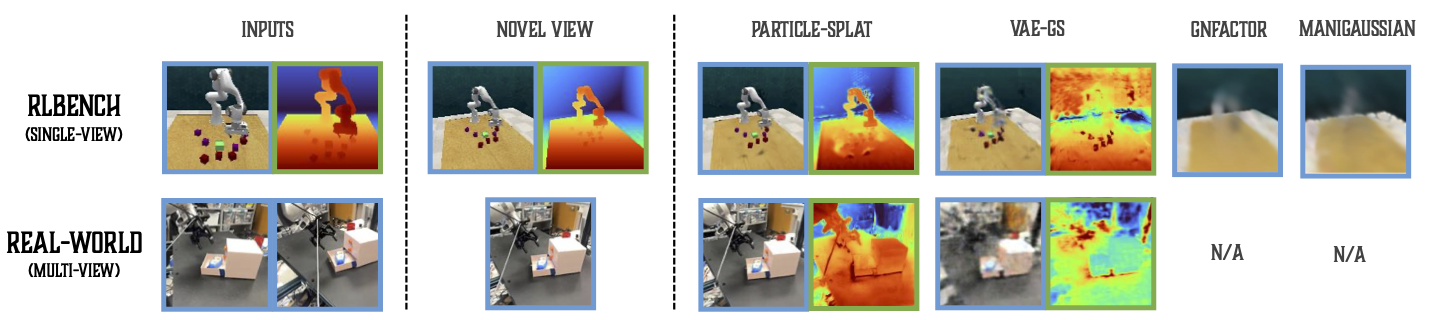}
    \caption{\textbf{Qualitative Novel View Synthesis Results.} We visualize novel view renderings produced by our learned representation on RLBench scenes (top) and real-world scenes (bottom), together with applicable results from baseline methods; zoom in for best visual comparison. Blue-framed images show RGB images, while green-framed images show depth images.}
    \label{fig:nvs_visuals}
    \vspace{-1em}
\end{figure}

\textbf{Results.}
Table~\ref{tab:nvs_results} reports quantitative novel view synthesis performance on RLBench and real-world scenes, while Fig.~\ref{fig:nvs_visuals} shows representative qualitative results. Both GNFactor~\cite{ze2023gnfactor} and ManiGaussian~\cite{lu2024manigaussian} are non-object-centric methods that rely on action-conditioned optimization, and they produce substantially lower-fidelity renderings as seen in Fig.~\ref{fig:nvs_visuals}, highlighting the challenge of learning high-quality multi-object scene representations when reconstruction and policy learning are jointly optimized. VAE-GS captures coarse texture and structure but lacks object-centric reasoning, leading to geometrically inconsistent Gaussian placements, structural misalignment, and inaccurate depth estimates (Fig.~\ref{fig:nvs_visuals}). In contrast, ParticleSplat achieves the best PSNR, SSIM, and LPIPS across both single-view and multi-view settings on RLBench, and maintains strong performance on real-world scenes from just two RGB inputs without depth supervision, where its rendered depth maps nevertheless remain geometrically consistent across views. Qualitatively, our method preserves sharp edges and high-frequency details, indicating that the object-centric particle representation simultaneously captures coherent 3D geometry and semantic consistency. Extended qualitative results, including larger reference-view gaps and depth visualizations, are provided in Appendix~\ref{subsec:nvs-extended}.

\textbf{Ablations.} We conduct ablations on novel view synthesis performance to validate key design choices. Full results and discussion are provided in Appendix~\ref{subsec:nvs-extended}. Removing the object-centric particle structure, replacing the Gaussian splatting decoder with a 2D pixel decoder, disabling pose conditioning via Plücker rays, and omitting depth feature fusion all substantially degrade the visual metrics, confirming the importance of each component.

\begin{table}[!t]
\centering
\scriptsize
\begingroup
\setlength{\tabcolsep}{3pt}
    \begin{tabular}{l|l|c|ccc}
        \toprule
        Dataset & Method & Input & PSNR $\uparrow$ & SSIM $\uparrow$ & LPIPS $\downarrow$ \\
        \midrule
        \multirow{8}{*}{RLBench} 
        & GNFactor~\cite{ze2023gnfactor}      & 1$\times$RGB-D & 16.20 & 0.5273 & 0.6819 \\
        & ManiGaussian~\cite{lu2024manigaussian}  & 1$\times$RGB-D & 15.87 & 0.5154 & 0.6793 \\
        & VAE-GS        & 1$\times$RGB-D & 21.85 & 0.6858 & 0.4838 \\
        & \textbf{ParticleSplat (Ours)}          & 1$\times$RGB-D & \textbf{25.35} & \textbf{0.7293} & \textbf{0.3563} \\
        \cmidrule(l){2-6}
        & VAE-GS        & 2$\times$RGB-D & 25.93 & 0.7326 & 0.3404 \\
        & \textbf{ParticleSplat (Ours)}          & 2$\times$RGB-D & \textbf{28.77} & \textbf{0.8144} & \textbf{0.2565} \\
        \midrule
        \multirow{2}{*}{Real-world}
        & VAE-GS        & 2$\times$RGB   & 20.33 & 0.4659 & 0.6587 \\
        & \textbf{ParticleSplat (Ours)}          & 2$\times$RGB   & \textbf{24.24} & \textbf{0.8455} & \textbf{0.1153} \\
        \bottomrule
    \end{tabular}
\endgroup
    \caption{\textbf{Quantitative novel view synthesis results.} We evaluate on RLBench (RGB-D) and real-world scenes (RGB-only), reporting PSNR$\uparrow$, SSIM$\uparrow$, and LPIPS$\downarrow$ under single-view (1$\times$RGB-D) and multi-view (2$\times$RGB-D / 2$\times$RGB) settings.}
    \label{tab:nvs_results}
\vspace{0.6em}
\begin{tabular}{l ccc c}
\toprule
\textbf{Method} / \textit{Task Group} & \begin{tabular}[c]{@{}c@{}}\textit{Object-centric}\\\textit{interaction}\end{tabular} & \begin{tabular}[c]{@{}c@{}}\textit{3D}\\\textit{geometry}\end{tabular} & \textit{Both} & \textbf{Overall} \\
\midrule
GNFactor~\cite{ze2023gnfactor} & 32.3 & 63.4 & 9.8 & 31.7 \\
ManiGaussian~\cite{lu2024manigaussian} & 52.3 & \underline{66.3} & 19.1 & 45.2 \\
VAE-GS & 48.9 & 62.8 & 16.8 & 42.0 \\
DLP~\cite{daniel2026lpwm} & \underline{64.7} & 37.0 & \underline{21.0} & \underline{46.6} \\
\textbf{ParticleSplat (Ours)} & \textbf{82.6} & \textbf{86.1} & \textbf{50.6} & \textbf{73.0} \\
\bottomrule
\end{tabular}
\caption{\textbf{Imitation learning performance with different representations (single-view).} Success rates (\%) on RLBench manipulation tasks using policies conditioned on different learned representations, averaged over the tasks of each task group defined in Sec.~\ref{subsec:exp_imit} and over all 10 tasks; per-task results are reported in Table~\ref{tab:manipulation_results_full} in Appendix~\ref{subsec:il-extended}. Results are means over three seeds (25 episodes per task per seed).}
\label{tab:manipulation_results}
\vspace{0.6em}
\begin{tabular}{l ccc c}
\toprule
\textbf{Method} / \textit{Task Group} & \textit{Basic} & \textit{Contact-Rich} & \textit{Long-Horizon} & \textbf{Overall} \\
\midrule
DLP~\cite{daniel2026lpwm} & 46.4 & 50.6 & 3.4 & 34.1 \\
\textbf{ParticleSplat (Ours)} & \textbf{80.4} & \textbf{66.9} & \textbf{27.9} & \textbf{56.1} \\
\bottomrule
\end{tabular}
\caption{\textbf{Additional imitation learning performance on MimicGen (multi-view).} Success rates (\%) on MimicGen~\cite{mandlekar2023mimicgen} manipulation tasks, averaged over the tasks of each official task group and over all 12 tasks. Per-task results, together with a comparison against the state-of-the-art methods on this benchmark, are reported in Table~\ref{tab:mimicgen_full} in Appendix~\ref{subsec:il-extended}. Results are means over three seeds (50 rollouts per task per seed). Multi-view DLP results are reproduced from~\cite{zhang20263ddlp}.}
\label{tab:mimicgen}
\vskip -0.12in
\end{table}

\subsection{Self-supervised Object-centric 3D Scene Decomposition}
\label{subsec:exp_ctrl}
\begin{figure}
    \centering
    \includegraphics[width=\linewidth]{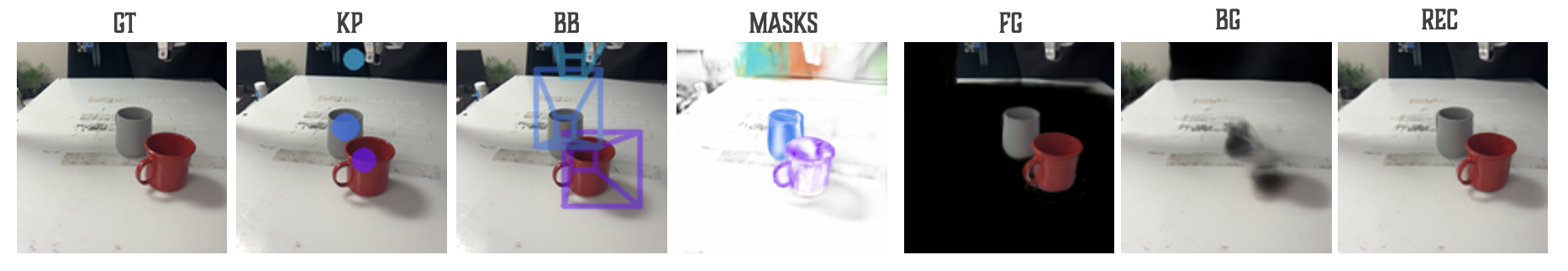}
    \caption{\textbf{Scene decomposition with ParticleSplat.} Self-supervised object-centric decomposition on real-world data.}
    \label{fig:latent_decomp}
    \vskip -0.12in
\end{figure}

\begin{figure}
    \centering
    \includegraphics[width=\linewidth]{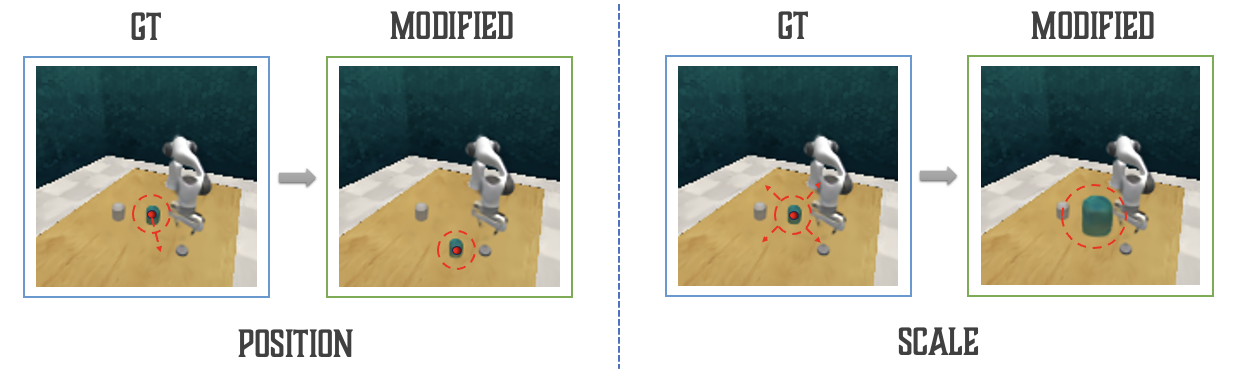}
    \caption{\textbf{Latent controllability.} Perturbing particle position $z_p$ translates objects in 3D space; scaling $z_s$ proportionally expands objects in their local frame.}
    \label{fig:latent_control}
    \vskip -0.12in
    \vspace{-1em}
\end{figure}

We showcase the resulting object-centric decomposition in Figures~\ref{fig:arch} and~\ref{fig:latent_decomp}. Since our self-supervised particles capture scene structure at a granularity that may differ from human-annotated masks, there is no unique ground-truth target for quantitatively evaluating objectness or latent edits. In particular, human masks are not always available, and when they are, they reflect semantic notions of objects that may not coincide with the scene parts induced by our reconstruction objective. Consequently, keypoint- and particle-based methods~\cite{kulkarni2019transporter,minderer2019structvrnn,daniel22adlp,qi2025ecdiffuser}, which ParticleSplat extends, evaluate discovered entities through their downstream use rather than against annotation masks. Following the same protocol, we use these visualizations, followed by downstream policy performance, as evidence of coherent and controllable entities. Notably, our method produces explicit, disentangled entities reflecting task-relevant semantics in Fig.~\ref{fig:latent_decomp}. We further demonstrate the interpretability and controllability of our latent space by manually perturbing particle attributes and observing the corresponding geometric changes in rendered scenes. As shown in Fig.~\ref{fig:latent_control}, we apply (i) translations to a particle's latent position $z_p$ and (ii) scaling to its latent scale $z_s$. Because each particle corresponds to a fixed set of aligned 3D Gaussians, these perturbations propagate consistently in an object-centric manner: translating $z_p$ rigidly shifts the object in 3D world space, while scaling $z_s$ expands the object proportionally in its local frame. These results highlight the structured, semantically meaningful nature of our particle latents, enabling intuitive 3D scene editing directly in the latent space.

\subsection{Imitation Learning with ParticleSplat}
\label{subsec:exp_imit}
We assess the benefits of our representation for downstream decision making via imitation learning. Our main benchmark is language-conditioned RLBench~\cite{james2020rlbench}, on which we compare and ablate all representations (Table~\ref{tab:manipulation_results}; per-task results in Table~\ref{tab:manipulation_results_full}). As a secondary benchmark, we use MimicGen~\cite{mandlekar2023mimicgen}, on which we compare with DLP and, in the appendix, with the state-of-the-art methods of that benchmark (Table~\ref{tab:mimicgen}; per-task results in Table~\ref{tab:mimicgen_full}). We reserve this section for RLBench and refer to Appendix~\ref{subsec:il-extended} for the MimicGen experimental setup and results analysis.

\textbf{Policies.} A single policy network cannot be shared across representation families. ManiGaussian~\cite{lu2024manigaussian} and GNFactor~\cite{ze2023gnfactor} output dense $100^3$ voxel volumes, which require an efficient-attention architecture (Perceiver~\cite{jaegle2021perceivergeneralperceptioniterative}) to learn a policy over the voxel lattice. The compact entity tokens of ParticleSplat, VAE-GS, and DLP instead admit a full-attention policy. Therefore, we use EC-Diffuser~\cite{qi2025ecdiffuser}, a diffusion-based behavioral cloning method that operates on entity tokens and models multi-object dynamics and actions in a permutation-equivariant manner. We adapt it to our multi-view latent particles by treating language instructions as additional ``particles'' and enabling joint self-attention between language and entity tokens (Appendix~\ref{subsec:ecdiff-lang}). Since neither policy applies to the other family, VAE-GS and DLP provide the comparison in which only the representation varies, while ManiGaussian and GNFactor serve as imitation-learning baselines whose results reflect representation and policy as a whole.

\textbf{Evaluation.}
On RLBench, we report task success rates averaged over three fixed random seeds, evaluating 25 episodes per task per seed, and summarize results as mean success and standard deviation across seeds. During evaluation, policies receive only a single RGB image and depth map from the front camera as input, and actions are generated by rolling out the learned policy. Additional implementation details are provided in Appendix~\ref{subsec:il-extended}.

\textbf{Results.}
Table~\ref{tab:manipulation_results} presents success rates on RLBench manipulation tasks (per-task results in Table~\ref{tab:manipulation_results_full} in the Appendix), categorized by reasoning requirements: (i) object-centric interaction tasks (\textit{push buttons}, \textit{meat off grill}, \textit{slide block}, \textit{drag stick}, \textit{sweep to dustpan}), (ii) 3D geometry tasks (\textit{turn tap}, \textit{open drawer}), and (iii) tasks requiring both (\textit{put in drawer}, \textit{stack blocks}, \textit{close jar}).

Object-centric interaction tasks primarily require entity tracking and interaction modeling, where 2D object-centric methods like DLP perform competitively. Geometry-focused tasks demand precise 3D pose estimation and occlusion reasoning, favoring reconstruction methods like ManiGaussian.
However, \emph{all baselines} struggle on tasks requiring \emph{both} object-centric reasoning \emph{and} accurate 3D structure---such as \textit{put in drawer}, \textit{stack blocks}, and \textit{close jar}---which involve long-horizon multi-object interactions with precise spatial alignment and occlusion handling. By jointly modeling object-centric structure and explicit 3D geometry, ParticleSplat achieves robust performance across all categories and substantial gains on these most challenging tasks. Overall, our method significantly outperforms all baselines (73.0\% vs.\ 31.7--46.6\%), confirming the value of 3D object-centric representations for robotic manipulation. The same trend holds on MimicGen (Table~\ref{tab:mimicgen}), where ParticleSplat improves over multi-view DLP on all three task groups (56.1\% vs.\ 34.1\% overall). Extended multi-view RLBench results and the full MimicGen results are in Appendix~\ref{subsec:il-extended}.

\section{Conclusion}
\label{sec:conc}

ParticleSplat introduces the first self-supervised object-centric representation trained end-to-end with feedforward 3D Gaussian splatting, bridging structured 2D latent particles with explicit 3D scene geometry. We demonstrate that this approach achieves strong novel view synthesis quality while providing an interpretable, controllable 3D latent space that enables scene editing and significantly improves performance on complex robotic manipulation tasks compared to non-object-centric 3D representations and 2D object-centric baselines.

\textbf{Limitations.} Like other self-supervised object-centric methods~\cite{locatello2020slotattn, daniel2026lpwm}, ParticleSplat performs best on scenes with limited background variability; scaling to large-scale in-the-wild environments and larger models remains an important challenge. The fixed particle budget $L$ also bounds the number of entities that can be represented, and entities beyond the budget are absorbed in the background. Additionally, our approach requires ground-truth camera poses during both training and inference. Future work could explore unposed multi-view settings, joint camera pose estimation, and dynamics prediction for world models~\cite{daniel2026lpwm}, and integration with reinforcement learning beyond imitation learning.

\newpage

{
    \small
    \bibliographystyle{ieeenat_fullname}
    \bibliography{main}
}

\clearpage
\newpage
\onecolumn
\appendix
\section{Appendix}

\subsection{ParticleSplat: Extended Method}
\label{subsec:apndx_method}
We seek to learn a structured 3D latent representation from a set of $V$ posed 2D observations $\mathcal{O} = \{ (\mathbf{x}^{v}, \mathbf{P}^{v}, \mathbf{K}^{v}) \}_{v=1}^{V}$. For each view $v$, the camera extrinsic pose is $\mathbf{P}^v = [\mathbf{R}^v | \mathbf{t}^v] \in SE(3)$, with rotation $\mathbf{R}^v \in SO(3)$ and translation $\mathbf{t}^v \in \mathbb{R}^3$. The camera intrinsics are given by $\mathbf{K}^v \in \mathbb{R}^{3 \times 3}$. Each observation $\mathbf{x}^{v} \in \mathbb{R}^{C \times H \times W}$ has spatial dimensions $H, W$ and $C \in \{3, 4\}$ channels for RGB or RGB-D inputs, respectively. This formulation supports both multi-view and monocular setups, with monocular RGB-D ($V=1$, $C=4$) strictly required to resolve scale ambiguity where object size and depth are indistinguishable without parallax.

Crucially, the particle structure in Deep Latent Particles (DLP)—with explicit position, scale, depth, transparency, and appearance attributes—closely mirrors the attributes of 3D Gaussian primitives in 3DGS, but remains confined to 2D image space. We bridge this gap by extending DLP particles to 3D scene space, enabling self-supervised learning of geometry-aware, object-centric representations that naturally yield segmentation masks without explicit supervision. This formulation adopts a self-supervised VAE with particle-based latents, where our view-aware encoder fuses 2D observations with camera geometry to propose particle posteriors, and our generative process employs a Gaussian Splatting decoder that transforms particles into 3D Gaussians via differentiable rasterization. We optimize a view-aware geometric evidence lower bound (ELBO) that balances reconstruction with view-dependent KL regularization. Together, this yields the first self-supervised object-centric VAE for 3D Gaussian Splatting, producing compact particle representations that enable both high-fidelity novel view synthesis and interpretable control for downstream decision-making. Our model is illustrated in Figure~\ref{fig:arch}.

\subsubsection{Multi-view Latent Particles}
\label{sec:method_particles}
We structure the latent space as a set of multi-view particles $\mathcal{Z}=
\{z_{\mathrm{fg}}^{v,\ell}\}_{v=1,\ell=1}^{V,L}
\;\cup\;
\{z_{\mathrm{bg}}^{v}\}_{v=1}^{V}$ consisting of \(V \times (L + 1)\) entities, where \(V\) is the number of views, \(L\) is the number of foreground particles, and each view is augmented with a single background particle. Each foreground particle $z_{\text{fg}} = [z_p, z_s, z_d, z_e, z_t, z_f] \in \mathbb{R}^{7 + d_{\text{obj}}}$ encodes disentangled stochastic attributes that explicitly separate \emph{image-plane} geometry (projected 2D appearance in the camera view) from \emph{volumetric} structure (true 3D extent in world coordinates). The image-plane attributes include position $z_p \sim \mathcal{N}(\mu_p, \sigma_p^2) \in \mathbb{R}^2$, representing 2D keypoint coordinates, and scale $z_s \sim \mathcal{N}(\mu_s, \sigma_s^2) \in \mathbb{R}^2$, defining the particle's spatial extent in pixel space (bounding-box dimensions). The volumetric attributes comprise depth $z_d \sim \mathcal{N}(\mu_d, \sigma_d^2) \in \mathbb{R}$, encoding distance from the camera along the viewing ray to determine compositing order, and depth extent $z_e \sim \mathcal{N}(\mu_e, \sigma_e^2) \in \mathbb{R}$, modeling the particle's thickness along the viewing direction. For example, a large $z_e$ creates ``deep'' objects spanning multiple depth planes (like a box), while a small $z_e$ yields ``flat'' objects concentrated near $z_d$ (like a sheet).
Each particle also includes transparency $z_t \sim \mathrm{Beta}(a,b) \in [0,1]$ for alpha compositing and visual features $z_f \sim \mathcal{N}(\mu_f, \sigma_f^2) \in \mathbb{R}^{d_{\text{obj}}}$ encoding local appearance and semantics. Per-view background particles $z_{\text{bg}} \sim \mathcal{N}(\mu_{\text{bg}}, \sigma_{\text{bg}}^2) \in \mathbb{R}^{d_{\text{bg}}}$ capture residual scene appearance not modeled by foreground particles. This factorization enables consistent lifting of 2D-inferred particles into coherent 3D representations.

\subsubsection{Encoder}
\label{sec:method_encoder}
The encoder $q_{\phi}(\mathcal{Z} \mid \mathcal{O})$ models the posterior distribution over latent particles given sets of posed observations. We extend the DLP encoder~\cite{daniel2026lpwm} to support multi-view inputs, depth, and camera geometry. The encoding proceeds hierarchically: keypoint proposals are first extracted from each view via differentiable spatial softmax (SSM~\cite{jakab2018unsupervised}) over patch-based feature maps. For each proposal, a spatial glimpse of dimensions $H_p \times W_p$ is extracted using a Spatial Transformer Network (STN~\cite{jaderberg2015stn}) with bilinear sampling, and the remaining particle attributes (Sec.~\ref{sec:method_particles}) are encoded via small CNNs. An attention-based interaction module~\cite{daniel2026lpwm} then refines relative depth and appearance features through cross-view attention over the particles of all context views. Extracting 3D geometry from 2D observations is ill-posed~\cite{longuet1981computer}, so we introduce camera-conditioned modifications detailed below.

\textbf{Geometric Priors via Ray Embeddings.} To inject 3D-aware priors, we fuse spatially-aligned geometric conditioning into both convolutional and attention modules. For each pixel at 2D coordinates $(u,v)$, we compute its corresponding camera ray with origin $\mathbf{o} \in \mathbb{R}^3$ (camera center from extrinsics) and direction $\mathbf{d} \in \mathbb{R}^3$ (unprojected via $\mathbf{K}^{-1}$, rotated by $\mathbf{R}$, and $L_2$-normalized):
\[
\mathbf{d} = \frac{\mathbf{RK}^{-1}[u, v, 1]^T}{\|\mathbf{RK}^{-1}[u, v, 1]^T\|}.
\]
We encode rays using Plücker coordinates $\mathbf{r} = (\mathbf{d}, \mathbf{o} \times \mathbf{d}) \in \mathbb{R}^6$~\cite{zhang2024cameras}, providing a continuous, translation-invariant parameterization of 3D lines.

\textbf{Convolutional Conditioning.} Let $\mathbf{h}_l \in \mathbb{R}^{C_l \times H_l \times W_l}$ be an intermediate feature map at layer $l$, and let $\mathcal{C} = \{\mathbf{c}_i\}_{i=1}^M$ be the set of $M$ geometric tensors (e.g., ray maps, optional depth). Each $\mathbf{c}_i$ is resampled to $(H_l, W_l)$ via bilinear interpolation, with direction components $L_2$-normalized post-interpolation. The tensors are then projected to channel space via $1\times1$ convolutions and additively fused:
\[
\mathbf{h}_l' = \mathbf{h}_l + \sum_{i=1}^M \left( \mathbf{W}_i * \phi(\tilde{\mathbf{c}}_i) + \mathbf{b}_i \right),
\]
where $\phi$ is an activation, and $\mathbf{W}_i, \mathbf{b}_i$ are learned per-tensor projections.

\textbf{Attention Conditioning.} For particle $z$ with image-plane position $z_p = (u_p, v_p)$, we compute its associated ray $\mathbf{r} \in \mathbb{R}^6$ as above. This pose signal modulates the particle embedding $\mathbf{h}_z$ via feature-wise linear modulation (FiLM~\cite{perez2018film}) before self-attention:
\[
\mathbf{h}_z = f(\mathbf{z}), \qquad (\boldsymbol{\gamma}, \boldsymbol{\beta}) = g(\mathbf{r}),
\]
\[
\tilde{\mathbf{h}}_z = \boldsymbol{\gamma} \odot \mathbf{h}_z + \boldsymbol{\beta},
\]
where $f$ projects to transformer dimension, $g$ maps rays to FiLM parameters (initialized near identity: $\boldsymbol{\gamma} \approx \mathbf{1}$, $\boldsymbol{\beta} \approx \mathbf{0}$), and $\odot$ is element-wise multiplication. Figure~\ref{fig:attention_conditioning} shows a high-level, PyTorch-style sketch of this ray-conditioned attention block.

\begin{figure}[!ht]
    \centering
    \begin{lstlisting}[style=mypython, basicstyle=\ttfamily\scriptsize]
h_z = f(z)                           # embed particle z into token features
r   = plucker_ray(z_p, K, R, o)      # ray from particle image location z_p=(u,v);
                                      # K: intrinsics, R: rotation, o: camera center
gamma, beta = g(r)                   # FiLM scale and shift from ray embedding
h_z_tilde = gamma * h_z + beta       # geometry-conditioned particle token
H = SelfAttention({h_z_tilde})       # refine all particle tokens jointly
    \end{lstlisting}
    \caption{\textbf{Ray-conditioned attention (PyTorch-style pseudocode).} Each particle token is FiLM-modulated using the Plücker ray corresponding to its image-plane position before self-attention.}
    \label{fig:attention_conditioning}
\end{figure}

\subsubsection{Decoder}
\label{sec:method_decoder}
The decoder models the likelihood $p_{\theta}(\mathbf{x} \mid \mathcal{Z})$ and defines the generative mapping from the latent particle set $\mathcal{Z}$ to image observations. Given $V$ posed cameras $\{(P^v, K^v)\}_{v=1}^{V}$ and the multi-view latent particle set
\[
\mathcal{Z}
=
\{z_{\mathrm{fg}}^{v,\ell}\}_{v=1,\ell=1}^{V,L}
\;\cup\;
\{z_{\mathrm{bg}}^{v}\}_{v=1}^{V},
\]
with $L$ foreground particles per view, the decoder first transforms all particles into an explicit set of world-space 3D Gaussian primitives and then renders them via a differentiable 3D Gaussian splatting rasterizer. Concretely, for each view $v$ and foreground particle $\ell$ we decode a \emph{particle-aligned} local Gaussian field anchored to that particle, while each background particle produces a dense, pixel-aligned background Gaussian field. The final scene representation is obtained by concatenating all foreground and background Gaussians and rendering them into each camera view to obtain RGB and depth reconstructions $(\hat{\mathbf{I}}^{v},\hat{\mathbf{D}}^{v}) \in \mathbb{R}^{3\times H\times W} \times \mathbb{R}^{1\times H\times W}$ for all $v\in\{1,\dots,V\}$. In practice, both the foreground and background decoders are implemented as CNNs with dedicated Gaussian heads that predict SH color, position offsets, depth residuals, scales, orientations, and opacities from particle appearance features.

\paragraph{Step 1: Foreground particle $\rightarrow$ canonical patch of Gaussians.}
Each foreground particle
\[
z_{\mathrm{fg}}^{v,\ell}
=
\big[
z_{p}^{v,\ell},
z_{s}^{v,\ell},
z_{d}^{v,\ell},
z_{e}^{v,\ell},
z_{t}^{v,\ell},
z_{f}^{v,\ell}
\big]
\]
is parameterized by image-plane position $z_{p}^{v,\ell}\in\mathbb{R}^2$, image-plane scale $z_{s}^{v,\ell}\in\mathbb{R}^2$, depth $z_{d}^{v,\ell}\in\mathbb{R}$, depth extent $z_{e}^{v,\ell}\in\mathbb{R}$, transparency $z_{t}^{v,\ell}\in[0,1]$, and an appearance latent $z_{f}^{v,\ell}\in\mathbb{R}^{d_{\mathrm{obj}}}$. The foreground decoder $D_{\mathrm{fg}}$ uses only the appearance latent to produce a fixed-resolution canonical patch of size $S_h \times S_w$, yielding a grid-aligned set of $N = S_h S_w$ Gaussians in a local patch coordinate frame:
\[
\mathscr{G}^{v,\ell}_{\mathrm{fg}}
=
D_{\mathrm{fg}}(z_{f}^{v,\ell})
=
\{
\mathbf{c}_{i}^{v,\ell},
\boldsymbol{\delta}_{i}^{v,\ell},
\Delta d_{i}^{v,\ell},
\mathbf{q}_{i}^{v,\ell},
\mathbf{s}_{i}^{v,\ell},
o_{i}^{v,\ell}
\}_{i=1}^{N}.
\]
Each Gaussian in this canonical patch is parameterized by spherical-harmonic (SH) color coefficients $\mathbf{c}_{i}^{v,\ell}$, a sub-cell offset $\boldsymbol{\delta}_{i}^{v,\ell}\in\mathbb{R}^2$, a depth residual $\Delta d_{i}^{v,\ell}\in\mathbb{R}$, a quaternion $\mathbf{q}_{i}^{v,\ell}\in\mathbb{R}^4$, anisotropic scales $\mathbf{s}_{i}^{v,\ell}\in\mathbb{R}^3$, and an opacity logit $o_{i}^{v,\ell}\in\mathbb{R}$. At this stage, all Gaussians live in a canonical patch coordinate system, analogous to the canonical glimpse frame in DLP.

\paragraph{Step 2: Canonical patch $\rightarrow$ image-plane Gaussians.}
In DLP, decoded RGB patches are “unwarped’’ back to the image using the STN sampling transform. Here, instead of resampling pixels, we must transform the \emph{Gaussian parameters} (centers and covariances) from patch coordinates into the full-image coordinate frame using the same affine map.

We represent the canonical patch grid by $\mathbf{u}\in[0,1]^2$ (one coordinate per cell). For each Gaussian, the decoder predicts a bounded sub-cell offset from $\boldsymbol{\delta}\in\mathbb{R}^2$ via a sigmoid:
\[
\Delta \mathbf{u}
=
\big(\sigma(\boldsymbol{\delta})-\tfrac12\big)\odot\big(\tfrac{1}{S_w},\tfrac{1}{S_h}\big),
\]
and the patch-normalized center is
\[
\mathbf{u}^{(\mathrm{patch})}
=
2(\mathbf{u}+\Delta \mathbf{u})-1,
\]
where the factor $2(\cdot)-1$ converts from $[0,1]^2$ to the STN convention $[-1,1]^2$.

The particle’s STN transform defines an affine map from patch-normalized to image-normalized coordinates:
\[
\begin{aligned}
\mathbf{A}^{v,\ell}
&=
\begin{pmatrix}
s_x & 0 & p_x\\
0 & s_y & p_y
\end{pmatrix},\\
(p_x,p_y) = z_p&^{v,\ell}, \qquad
(s_x,s_y) = \sigma(z_s^{v,\ell}).
\end{aligned}
\]
Lifting to homogeneous coordinates
\(
\mathbf{v}^{(\mathrm{patch})}
=
[\mathbf{u}^{(\mathrm{patch})}; 1]
\),
the corresponding image-frame centers in $[-1,1]^2$ are
\[
\mathbf{u}^{(\mathrm{img})} = \mathbf{A}^{v,\ell}\,\mathbf{v}^{(\mathrm{patch})},
\]
and we finally map to pixel-normalized coordinates $[0,1]^2$ via
\[
\mathbf{u}^{(\mathrm{img})} \leftarrow \tfrac{1}{2}(\mathbf{u}^{(\mathrm{img})}+1).
\]

We apply an analogous transformation to Gaussian covariances. Let $\Sigma^{(\mathrm{patch})}\in\mathbb{R}^{3\times 3}$ be a covariance in the canonical patch frame. The Jacobian $\mathbf{J}^{v,\ell}$ of the affine map scales the x,y axes according to the particle scale and patch size,
\[
\mathbf{J}^{v,\ell}
=
\mathrm{diag}\!\left(\tfrac{s_x}{s_w},\,\tfrac{s_y}{s_h},\,1\right),
\qquad
s_w = S_w/W,\;\; s_h = S_h/H,
\]
and the corresponding covariance in the image-aligned frame is
\[
\Sigma^{(\mathrm{img})}
=
\mathbf{J}^{v,\ell}\,\Sigma^{(\mathrm{patch})}\,(\mathbf{J}^{v,\ell})^{\top}.
\]
This yields per-Gaussian centers and covariances that are properly placed and scaled in the full image domain. Non-spatial attributes (SH color, depth residuals, opacity logits) are carried over unchanged at this stage.

\paragraph{Step 3: Image-plane Gaussians $\rightarrow$ world-space 3D Gaussians.}
The latent particle attributes $(z_d^{v,\ell}, z_e^{v,\ell}, z_t^{v,\ell})$, together with camera geometry, provide exactly the degrees of freedom needed to place Gaussians in 3D. For each view $v$ and particle $\ell$, we transform the image-plane Gaussians
\[
\mathscr{G}^{v,\ell}_{\mathrm{fg}}
=
\{\mathbf{c}^{v,\ell}_i,\boldsymbol{\delta}^{v,\ell}_i,\Delta d^{v,\ell}_i,\mathbf{q}^{v,\ell}_i,\mathbf{s}^{v,\ell}_i,o^{v,\ell}_i\}_{i=1}^{N}
\]
into a set of world-space 3D Gaussians
\[
\mathcal{G}^{v,\ell}_{\mathrm{fg}}
=
\{(\boldsymbol{\mu}^{v,\ell}_i,\Sigma^{v,\ell}_i,\mathbf{c}^{v,\ell}_i,\alpha^{v,\ell}_i)\}_{i=1}^{N},
\]
where $\boldsymbol{\mu}^{v,\ell}_i\in\mathbb{R}^3$ and $\Sigma^{v,\ell}_i\in\mathbb{R}^{3\times 3}$ are the world-space mean and covariance, $\mathbf{c}^{v,\ell}_i$ are SH color coefficients, and $\alpha^{v,\ell}_i\in[0,1]$ are the final opacities.

\emph{Depth and 3D means.}
We first map the particle depth and extent latents to a bounded metric depth along the camera $Z$ axis:
\[
\begin{aligned}
d^{v,\ell} &= d_{\min} + \sigma(z_{d}^{v,\ell})\,(d_{\max}-d_{\min}),\\
\Delta^{v,\ell} &= \tfrac{1}{2}(d_{\max}-d_{\min})\,\sigma(z_{e}^{v,\ell}),\\
\zeta^{v,\ell}_i &= d^{v,\ell} + \Delta^{v,\ell}\tanh(\Delta d^{v,\ell}_i),
\end{aligned}
\]
where $[d_{\min}, d_{\max}]$ defines the valid depth range and $\zeta^{v,\ell}_i$ is the planar-$Z$ depth for Gaussian $i$. Let $\mathbf{u}^{v,\ell}_i\in[0,1]^2$ be its pixel-normalized image-plane center. Using the camera pose $P^v=[\mathbf{R}^v\,|\,\mathbf{o}^v]$ and intrinsics $\mathbf{K}^v$, we form an unnormalized ray direction
\[
\mathbf{o}^v \in \mathbb{R}^3,\qquad 
\mathbf{d}^v(\mathbf{u}) = \mathbf{R}^v(\mathbf{K}^v)^{-1}
\begin{bmatrix}u_x\\u_y\\1\end{bmatrix}.
\]
The world-space mean is then
\[
\boldsymbol{\mu}^{v,\ell}_i
=
\mathbf{o}^v + \zeta^{v,\ell}_i\,\mathbf{d}^v(\mathbf{u}^{v,\ell}_i).
\]

\emph{Covariances.}
We post-process predicted scales and quaternions to enforce validity:
\[
\overline{\mathbf{s}}^{v,\ell}_i
=
\mathbf{s}_{\min} + (\mathbf{s}_{\max}-\mathbf{s}_{\min})\,\sigma(\mathbf{s}^{v,\ell}_i),
\qquad
\overline{\mathbf{q}}^{v,\ell}_i
=
\frac{\mathbf{q}^{v,\ell}_i}{\|\mathbf{q}^{v,\ell}_i\|+\epsilon}.
\]
A positive semi-definite covariance in the camera-aligned local frame is constructed as
\[
\Sigma^{v,\ell}_{i,\mathrm{cam}}
=
\mathbf{Q}(\overline{\mathbf{q}}^{v,\ell}_i)\,
\mathrm{diag}\!\big(\overline{\mathbf{s}}^{v,\ell}_i \odot \overline{\mathbf{s}}^{v,\ell}_i\big)\,
\mathbf{Q}(\overline{\mathbf{q}}^{v,\ell}_i)^\top,
\]
where $\mathbf{Q}(\cdot)$ converts a unit quaternion to a $3\times 3$ rotation matrix. We then rotate covariances into the world frame using $\mathbf{R}^v$:
\[
\Sigma^{v,\ell}_i
=
\mathbf{R}^v\,\Sigma^{v,\ell}_{i,\mathrm{cam}}\,(\mathbf{R}^v)^\top.
\]

\emph{Opacity and SH coefficients.}
We map opacity logits to bounded opacities and gate them by the particle transparency,
\[
\alpha^{v,\ell}_i
=
z_t^{v,\ell}\;\psi\!\big(\sigma(o^{v,\ell}_i)\big),
\]
where $\psi(\cdot)$ is a monotonic opacity mapping (optionally scheduled during warm-up). The decoded SH coefficients $\mathbf{c}^{v,\ell}_i$ parameterize view-dependent appearance; we rotate them into a consistent world SH frame using $\mathbf{R}^v$ (standard SH rotation). Altogether, each foreground particle yields a set of world-space 3D Gaussians $\mathcal{G}^{v,\ell}_{\mathrm{fg}}$ fully determined by its attributes and the camera geometry.

\paragraph{Background particle $\rightarrow$ world-space Gaussians.}
Each view $v$ additionally contains a single background latent particle $z_{\mathrm{bg}}^{v}$, decoded by a global background decoder $D_{\mathrm{bg}}$ into a dense $H\times W$ field of pixel-aligned Gaussians:
\[
\mathscr{G}^{v}_{\mathrm{bg}}
=
D_{\mathrm{bg}}(z_{\mathrm{bg}}^{v}).
\]
Since these Gaussians are already defined at full-image resolution, they do not require the patch-to-image affine unwarping step used for foreground glimpses. We lift each background Gaussian to world space using the same camera-geometry mapping (ray and depth parameterization) as above, obtaining $\mathcal{G}^{v}_{\mathrm{bg}}$.

\paragraph{Stitching and differentiable rendering.}
We form global foreground and background Gaussian sets by aggregating across views and particles:
\[
\mathcal{G}_{\mathrm{fg}} = \bigcup_{v,\ell} \mathcal{G}^{v,\ell}_{\mathrm{fg}},
\qquad
\mathcal{G}_{\mathrm{bg}} = \bigcup_{v} \mathcal{G}^{v}_{\mathrm{bg}},
\qquad
\mathcal{G} = \mathcal{G}_{\mathrm{fg}} \cup \mathcal{G}_{\mathrm{bg}}.
\]
Thanks to the compositional nature of Gaussian splatting, stitching is implemented as simple concatenation. Given any target camera $(P^{t},K^{t})$, we render RGB and depth using the standard 3DGS rasterizer with front-to-back alpha compositing,
\[
(\hat{\mathbf{I}}^{t}, \hat{\mathbf{D}}^{t}) = \mathcal{R}(\mathcal{G}; P^{t}, K^{t}),
\]
enabling multi-view reconstruction and novel view synthesis directly from the compact object-centric latent particles.

Moreover, we can render each foreground particle independently to obtain an explicit 3D object decomposition. Let $\mathcal{G}^{v,\ell}_{\mathrm{fg}}$ be the set of world-space Gaussians associated with particle $\ell$ from view $v$. For a target camera $(P^{t},K^{t})$, per-particle RGBD is obtained by
\[
(\hat{\mathbf{I}}^{t,v,\ell}_\mathrm{fg}, \hat{\mathbf{D}}^{t,v,\ell}_\mathrm{fg}) = \mathcal{R}(\mathcal{G}^{v,\ell}_{\mathrm{fg}}; P^{t}, K^{t}),
\]
while background-only RGBD is rendered as
\[
(\hat{\mathbf{I}}^{t}_\mathrm{bg}, \hat{\mathbf{D}}^{t}_\mathrm{bg}) = \mathcal{R}(\mathcal{G}_{\mathrm{bg}}; P^{t}, K^{t}).
\]
More generally, any subset $\mathcal{G}_{\mathrm{sub}}\subseteq\mathcal{G}$ can be rendered via
\[
(\hat{\mathbf{I}}^{t}_{\mathrm{sub}}, \hat{\mathbf{D}}^{t}_{\mathrm{sub}}) = \mathcal{R}(\mathcal{G}_{\mathrm{sub}}; P^{t}, K^{t}),
\]
supporting object-wise masks, scene editing, and controllable recomposition directly in 3D Gaussian space.

\subsubsection{Loss}
\label{sec:method_elbo}
Given a set of posed observations $\mathcal{O}=\{(\mathbf{x}^v,\mathbf{P}^v,\mathbf{K}^v)\}$, we partition the views into a context set $\mathcal{C}$ and a target set $\mathcal{T}$. The context views are encoded to infer the multi-view latent particle representation $\mathcal{Z}$, while the target views are used to supervise novel views and encourage 3D consistency and occlusion reasoning. The decoder renders per-view RGB and depth by transforming $\mathcal{Z}$ into 3D Gaussians and rasterizing them under the corresponding camera parameters. We modify the ELBO objective to account for view geometry as follows.

We treat the multi-view latent particles $\mathcal{Z}$ as describing a \emph{single} underlying 3D scene, while each observation $\mathbf{x}^v$ is a 2D projection of that scene. Reconstruction terms are averaged over all context and target views, whereas the KL term is applied once at the scene level and is not normalized by the number of views. The negative ELBO is
\begin{equation*}
\begin{aligned}
\mathcal{L}_{\text{ELBO}}
&=
\underbrace{\frac{1}{|\mathcal{C}|+|\mathcal{T}|}
\Big(
\beta_{\mathrm{rec}}^{\mathrm{ctx}}\!\!\sum_{v\in\mathcal{C}}\!\!
\mathcal{E}(\mathbf{x}^v, \hat{\mathbf{x}}^v)
+
\beta_{\mathrm{rec}}^{\mathrm{tgt}}\!\!\sum_{t\in\mathcal{T}}\!\!
\mathcal{E}(\mathbf{x}^t, \hat{\mathbf{x}}^t)
\Big)}_{\mathcal{L}_{\mathrm{rec}}}
\\
&\quad+
\beta_{\mathrm{KL}}
\underbrace{\mathrm{KL}\!\left(q_{\phi}(\mathcal{Z}\mid\mathcal{C})\,\Vert\,p(\mathcal{Z})\right)}_{\mathcal{L}_{\mathrm{KL}}},
\end{aligned}
\end{equation*}
where $\beta_{\mathrm{rec}}^{\mathrm{ctx}}$ and $\beta_{\mathrm{rec}}^{\mathrm{tgt}}$ weight reconstruction on context and target views, respectively, and $\beta_{\mathrm{KL}}$ controls the strength of scene-level regularization.

\paragraph{Reconstruction loss}
For each view $i$, the reconstruction compares the ground-truth RGB–depth pair $\mathbf{x}^{i}=(\mathbf{I}^{i},\mathbf{D}^{i})$ with its rendering $\hat{\mathbf{x}}^{i}=(\hat{\mathbf{I}}^{i},\hat{\mathbf{D}}^{i})$:
\begin{equation*}
\mathcal{E}(\mathbf{x}^{i}, \hat{\mathbf{x}}^{i})
=
\mathcal{L}_{\mathrm{img}}^{(i)}
+
\lambda_{D}^{(i)}\,\big\|\tilde{\mathbf{D}}^{i}-\tilde{\hat{\mathbf{D}}}^{i}\big\|_{2}^2,
\end{equation*}
where $\lambda_{D}^{(i)}$ is a scalar weight that balances the depth reconstruction term relative to the image loss for view $i$, and the image reconstruction term $\mathcal{L}_{\mathrm{img}}^{(i)}$ is either a pixel-wise mean-squared error,
\[
\mathcal{L}_{\mathrm{img}}^{(i)}=\big\|\mathbf{I}^{i}-\hat{\mathbf{I}}^{i}\big\|_{2}^{2},
\]
or a VGG-based perceptual loss~\cite{hoshen2019perceptualloss},
\[
\mathcal{L}_{\mathrm{img}}^{(i)}=
\sum_{\ell}
\big\|
\Phi_{\ell}(\mathbf{I}^{i})
-
\Phi_{\ell}(\hat{\mathbf{I}}^{i})
\big\|_{2}^{2},
\]
with $\Phi_{\ell}$ denoting features extracted from layer $\ell$ of a fixed pretrained VGG network.

Following Splatt3R~\cite{smart2024splatt3r}, we additionally introduce a supervision mask $M^{i}\in\{0,1\}^{H\times W}$ in the monocular setting ($|\mathcal{C}|=1$) to restrict losses to pixels that are geometrically reconstructible from the single context view. When rendering target views that may cover regions outside the context camera frustum, pixels with $M^{i}=0$ are excluded from both image and depth terms, while pixels with $M^{i}=1$ receive full supervision. This masking avoids penalizing unavoidable hallucinations and stabilizes training under wide-baseline novel-view supervision from monocular input.

\paragraph{KL-divergence loss.}
Following DLP~\cite{daniel2024ddlp}, we regularize the inferred latent particles with a masked KL-divergence against fixed priors. In our setting, however, the latent set $\mathcal{Z}$ represents a \emph{single 3D scene} inferred from the context set $\mathcal{C}$, rather than a sequence of per-frame latent states. Accordingly, the KL term is applied once at the scene level and is not averaged over views. Recall that
\[
\mathcal{Z}
=
\{z_{\mathrm{fg}}^{v,\ell}\}_{v=1,\ell=1}^{V,L}
\;\cup\;
\{z_{\mathrm{bg}}^{v}\}_{v=1}^{V},
\]
where each foreground particle is
\[
z_{\mathrm{fg}}^{v,\ell}
=
\big[
z_{p}^{v,\ell},
z_{s}^{v,\ell},
z_{d}^{v,\ell},
z_{e}^{v,\ell},
z_{t}^{v,\ell},
z_{f}^{v,\ell}
\big].
\]
Here, $z_{p}^{v,\ell}$ and $z_{s}^{v,\ell}$ denote image-plane offset and scale, $z_{d}^{v,\ell}$ and $z_{e}^{v,\ell}$ denote depth and depth extent, $z_{t}^{v,\ell}$ denotes transparency, and $z_{f}^{v,\ell}$ denotes the particle appearance feature. The background latent ${z_{\mathrm{bg}}^{v}}$ contains only background appearance features, following LPWM.

We adopt the masked KL formulation of DLP, where particles with low transparency contribute less to the KL penalty for geometric and appearance attributes. The scene-level KL loss is 
\begin{align*}
\mathcal{L}_{\mathrm{KL}}
&=
\sum_{v=0}^{V}
\sum_{\ell=0}^{L}
\Bigg(
\sum_{\mathrm{att}\in\{o,s,d,e\}}
\mathrm{KL}\!\left(
q_{\phi}(z_{\mathrm{att}}^{v,\ell}\mid \mathcal{C})
\,\Vert\,
p_{\mathrm{att}}(z)
\right)\odot z_t^{v,\ell}
\nonumber\\
&\qquad\qquad
+
\mathrm{KL}\!\left(
q_{\phi}(z_t^{v,\ell}\mid \mathcal{C})
\,\Vert\,
p_t(z)
\right)
+
\beta_f\,
\mathrm{KL}\!\left(
q_{\phi}(z_f^{v,\ell}\mid \mathcal{C})
\,\Vert\,
p_f(z)
\right)\odot z_t^{v,\ell}
\Bigg)
\nonumber\\
&\qquad
+
\sum_{v=0}^{V}
\beta_f\,
\mathrm{KL}\!\left(
q_{\phi}(z_{\mathrm{bg}}\mid \mathcal{C})
\,\Vert\,
p_{\mathrm{bg}}(z)
\right),
\label{eq:scene_kl}
\end{align*}
where $\beta_f$ balances the KL contribution of appearance features relative to geometric attributes. As in DLP, the KL terms for geometry and appearance are masked by $z_t^{v,\ell}$, while the KL term for transparency itself is not masked. Intuitively, particles that are effectively inactive should not be penalized for their geometric or appearance latents, but the model must still learn whether each particle should be active.

Relative to DLP, the key difference is that the posterior is conditioned on the multi-view context set $\mathcal{C}$ and parameterizes a single scene-level latent representation shared across all rendered views. Therefore, the KL divergence loss regularizes the scene decomposition jointly across observations, encouraging a compact particle representation whose projections remain consistent under changes in viewpoint.

\paragraph{Particle regularization.}
In addition to the KL term, we optionally follow DLP~\cite{daniel2024ddlp} and regularize particle transparency to discourage the degenerate solution in which all particles remain active. In our setting, this failure mode would allow the model to explain a scene using many redundant particles, weakening object-centric decomposition and reducing 3D consistency. We therefore apply an $L_2$ penalty to the transparency variables:
\begin{equation*}
\mathcal{L}_{\mathrm{reg}}
=
\sum_{v=0}^{V} \sum_{\ell=0}^{L}
(z_t^{v, \ell})^2.
\label{eq:particle_reg}
\end{equation*}
This regularization biases the model toward activating only a limited number of particles to explain the scene. As particles with near-zero transparency contribute little to the rendered output, the model is encouraged to allocate representational capacity to a smaller set of informative particles. This promotes a more efficient scene decomposition and reduces redundancy among particles, which in turn stabilizes training and improves the consistency of the decoded 3D Gaussian representation during novel-view rendering.

The final training objective augments the ELBO with this regularization term:
\begin{equation*}
\mathcal{L}_{\text{ELBO}}
=
\mathcal{L}_{\mathrm{rec}}
+
\beta_{\mathrm{KL}}\,\mathcal{L}_{\mathrm{KL}}
+
\beta_{\mathrm{reg}}\,\mathcal{L}_{\mathrm{reg}},
\label{eq:final_loss}
\end{equation*}
where $\beta_{\mathrm{reg}}$ controls the strength of particle sparsity regularization. Following DLP, we set $\beta_{\mathrm{reg}}=\beta_{\mathrm{KL}}$ unless stated otherwise. We provide the detailed hyperparameters for the prior distributions of different attributes in Table~\ref{tab:hyperparams_prior}.

\subsection{Experiments Details}
\label{subsec:apndx_rlbench}
\subsubsection{RLBench Experiments}

\begin{table}[t]
\centering
\scriptsize
\renewcommand{\arraystretch}{1.05}
\setlength{\tabcolsep}{8pt}

\begin{tabular}{@{}>{\ttfamily}l l c l@{}}
\toprule
\rmfamily Task & Variation Type & Number of Variations & Language Template \\
\midrule
close\_jar         & color        & 20 & ``close the \underline{\hspace{0.4cm}} jar'' \\
open\_drawer       & placement    & 3  & ``open the \underline{\hspace{0.4cm}} drawer'' \\
sweep\_to\_dustpan  & size         & 2  & ``sweep dirt to the \underline{\hspace{0.4cm}} dustpan'' \\
meat\_off\_grill    & category     & 2  & ``take the \underline{\hspace{0.4cm}} off the grill'' \\
turn\_tap          & placement    & 2  & ``turn \underline{\hspace{0.4cm}} tap'' \\
slide\_block       & color        & 4  & ``slide the block to \underline{\hspace{0.4cm}} target'' \\
put\_in\_drawer     & placement    & 3  & ``put the item in the \underline{\hspace{0.4cm}} drawer'' \\
drag\_stick        & color        & 20 & ``use the stick to drag the cube onto the \underline{\hspace{0.4cm}} \underline{\hspace{0.4cm}} target'' \\
push\_buttons      & color        & 50 & ``push the \underline{\hspace{0.4cm}} button, [then the \underline{\hspace{0.4cm}} button]'' \\
stack\_blocks      & color, count & 60 & ``stack \underline{\hspace{0.4cm}} \underline{\hspace{0.4cm}} blocks'' \\
\bottomrule
\end{tabular}
\caption{\textbf{RLBench dataset task variations and corresponding language templates.} Each task is described by its own variation type, number of variations, and associated language template.}
\label{tab:task_variations}

\end{table}

We evaluate our method on RLBench~\cite{james2020rlbench} in simulation, a widely used benchmark for language-conditioned robotic manipulation that provides multi-view visual observations and diverse task variations.

\paragraph{Dataset overview.}
To ensure fair comparison with prior work, we follow exactly the same 10-task subset (see Table~\ref{tab:task_variations}) used by GNFactor~\cite{ze2023gnfactor} and ManiGaussian~\cite{lu2024manigaussian}. This setup contains 166 task variations in total. During training, we use 20 demonstrations per task, resulting in 200 training episodes overall. During evaluation, we run 25 rollout episodes per task.

\paragraph{Task variations.}
The 166 variations arise from controlled changes in both object attributes and scene layout. In particular, RLBench varies object appearance over a palette of 20 colors, object extent over two size regimes (short and tall), and task cardinality over three count settings (1, 2, and 3 objects), while placement-specific and category-specific options are defined on a per-task basis. For example, tasks such as \texttt{open\_drawer} vary by the target drawer location, whereas other tasks vary the manipulated object category or its spatial target. Beyond these discrete semantic factors, the initial arrangement of objects is further randomized by sampling tabletop poses within bounded regions, introducing additional geometric diversity across demonstrations. Together, these sources of variation require the policy to learn transferable structure across tasks rather than overfitting to a narrow set of scene instantiations.

\paragraph{Language instructions.}
For each episode, RLBench provides a natural language instruction describing the manipulation goal. We use these language annotations as the task specification paired with the corresponding visual observations, and encode the language instructions using a pre-trained CLIP~\cite{radford2021clip} language encoder following prior works~\cite{ze2023gnfactor, lu2024manigaussian, shridhar2023perceiver}.

\paragraph{Camera views.}

\begin{figure*}[t]
    \centering
    \includegraphics[width=\linewidth]{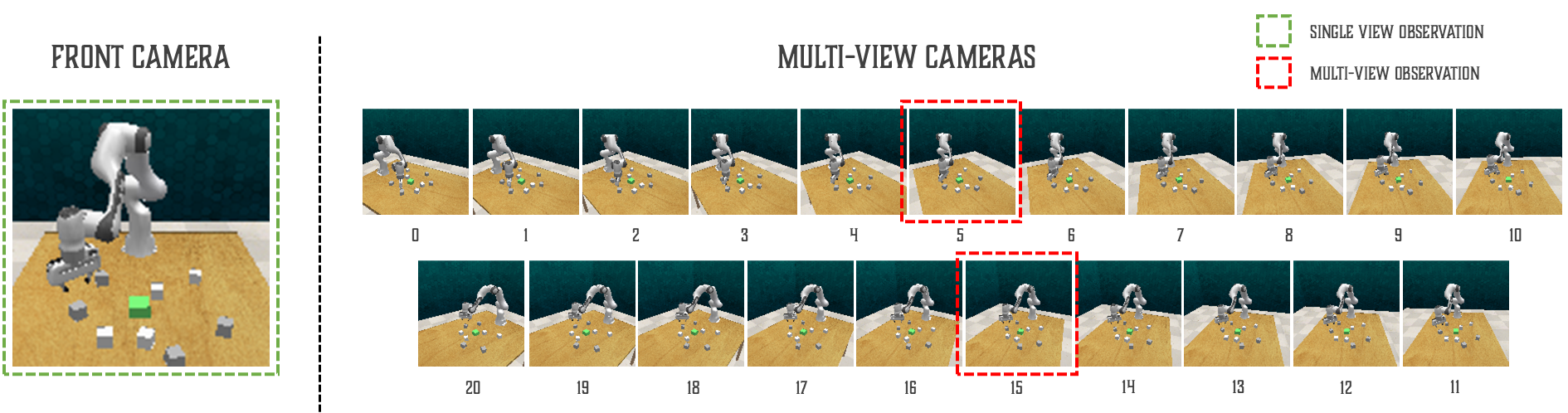}
    \caption{\textbf{RLBench camera views.}
    \textbf{Left:} the front camera observation.
    \textbf{Right:} the auxiliary 21-view camera set rendered around the scene.
    The \textcolor{green!60!black}{green box} marks the viewpoint used to train and evaluate the imitation learning policy in the \textit{single-view} setting.
    The \textcolor{red}{red boxes} indicate the viewpoints used for training and evaluation in the \textit{multi-view} setting.}
    \label{fig:rlbench_cameras}
\end{figure*}

RLBench provides synchronized observations from multiple cameras for each episode (as shown in Fig~\ref{fig:rlbench_cameras}). In our setup, we use the front camera observation as the primary input stream and retain its associated RGB image, depth map, and camera pose. In addition, following the data generation protocol of GNFactor and ManiGaussian, we render an auxiliary 21-view multi-view dataset for each scene, consisting of RGB images, depth maps, and corresponding camera poses. These auxiliary views provide richer geometric supervision for 3D representation learning.

\paragraph{3D reconstruction details.}
We train our representation using a 3D reconstruction objective, where the scene is reconstructed conditioned on a set of context cameras $\mathcal{C}$ and supervised using images from a set of target cameras $\mathcal{T}$. During training, the reconstruction is evaluated against the union of context and target views, i.e., $\mathcal{C} \cup \mathcal{T}$, while predictions are conditioned on the context observations.

For the multi-view setting, context and target views are sampled from the auxiliary set of 21 cameras rendered around the scene. Following the view sampling strategy of~\cite{chen2024mvsplat}, we sample two context views ($|\mathcal{C}|=2$) and four target views ($|\mathcal{T}|=4$). At the beginning of training, the two context views are selected to be spatially close to each other, with a \textit{view index gap} sampled from the range $[1,4]$. As training progresses, the separation between the context views is gradually increased, allowing the view index gap to expand to $[5,20]$, with the maximum gap reached halfway through the training schedule. Target views are then randomly sampled from the views whose indices lie between the two selected context views.

In the single-view setting, the front camera is used as the sole context view ($|\mathcal{C}|=1$), and two target views ($|\mathcal{T}|=2$) are sampled from the auxiliary set of 21 cameras. Since the context view is fixed, target views are sampled relative to the front camera. At the start of training, the target views are chosen from viewpoints close to the front camera, centered around the view index that most closely corresponds to the front observation (index 10), with a view index gap in the range $[1,2]$ (approximately indices 8--12). As training progresses, the sampling range gradually expands, increasing the view index gap to $[5,10]$, with the maximum gap reached halfway through training.

\begin{table}[t]
\centering
\scriptsize
\renewcommand{\arraystretch}{1.05}
\setlength{\tabcolsep}{8pt}

\begin{tabular}{@{}>{\ttfamily}l l c@{}}
\toprule
\rmfamily Task & Video Clip & Horizon (timesteps) \\
\midrule

\multirow{2}{*}{task\_a}
 & clip\_0 & 514 \\
 & clip\_1 & 427 \\
 & clip\_2 & 495 \\

\midrule

\multirow{6}{*}{antacid\_in\_drawer}
 & clip\_0 & 340 \\
 & clip\_1 & 279 \\
 & clip\_2 & 303 \\
 & clip\_3 & 278 \\
 & clip\_4 & 285 \\
 & clip\_5 & 286 \\

\midrule

\multirow{1}{*}{Total}
 & & 3207 \\

\bottomrule
\end{tabular}
\caption{\textbf{Real-world dataset video clips and planning horizons for each task.} Each task contains multiple video clips of different configurations with different time-step horizons.}
\label{tab:video_horizons}
\end{table}

\paragraph{Imitation learning details.}
Once the representation model is trained, we freeze its parameters and use the encoder to extract features from the demonstration observations. These encoded representations serve as inputs to our imitation learning policy, which is then trained to imitate the demonstrations of the RLbench manipulation tasks. The imitation learning policy is a modified version of EC-Diffuser~\cite{qi2025ecdiffuser} that predicts a sequence of actions with a planning horizon of 5 given the current encoded observation and additionally a language instruction conditioning. We include the details of its implementation in Sec.~\ref{subsec:ecdiff-lang}. At execution time, only the first action in the predicted trajectory is applied to the environment before the observation is updated and the next prediction step begins, following a receding-horizon control scheme. Under the multi-view setting, the policy receives observations from a fixed subset of cameras selected from the auxiliary 21-view camera set, as indicated by the \textcolor{red}{red boxes} in Fig.~\ref{fig:rlbench_cameras}. Under the single-view setting, the policy instead takes only the front camera observation as input, as indicated by the \textcolor{green!60!black}{green box} in Fig.~\ref{fig:rlbench_cameras}. Further implementation details of the imitation policy are provided in Sec.~\ref{subsec:ecdiff-lang}.

\begin{figure}[t]
    \centering
    \includegraphics[width=\linewidth]{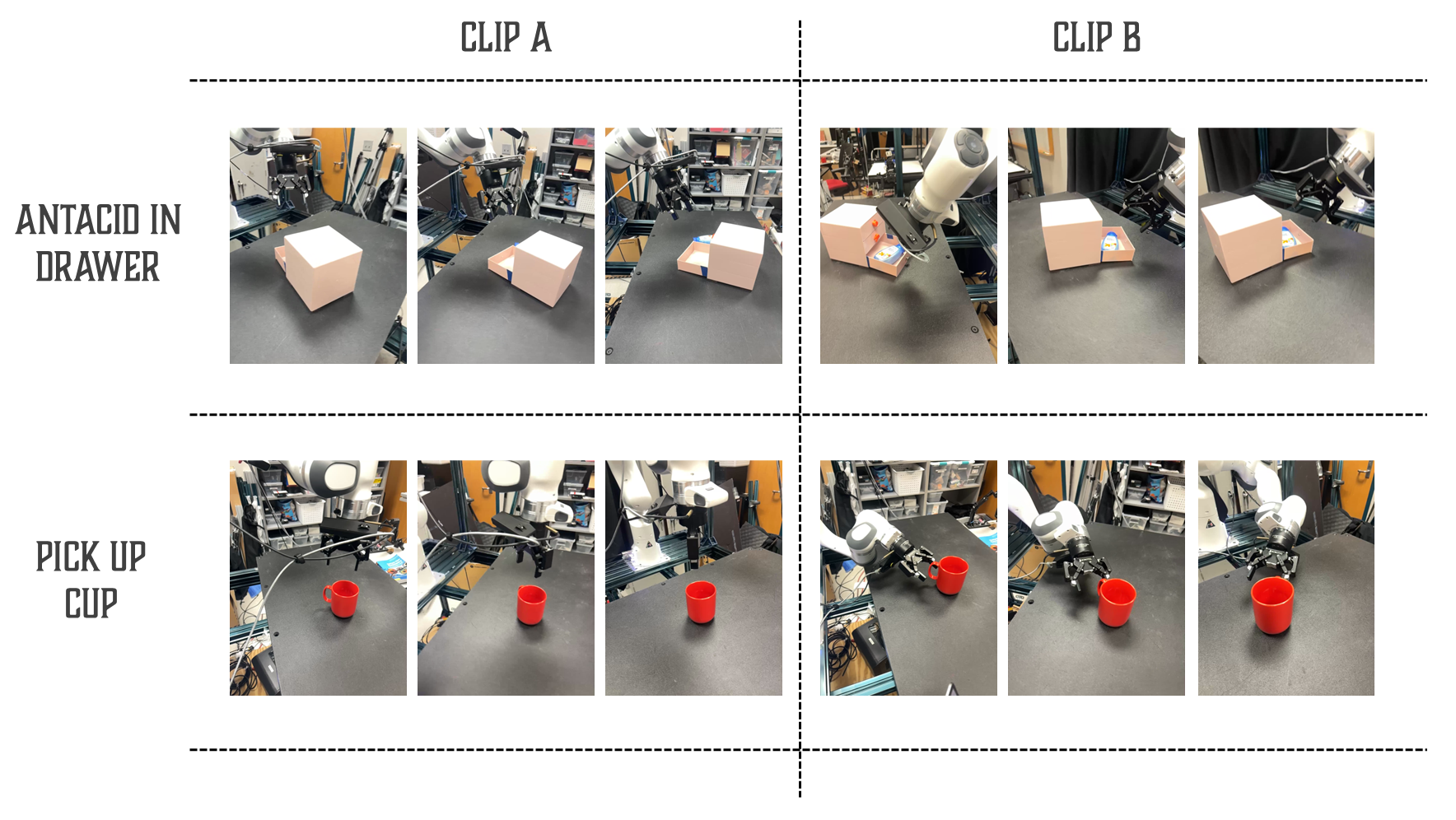}
    \caption{\textbf{Samples from the Real-world dataset.} \textbf{Top: } selected RGB images from video clips of \texttt{antacid\_in\_drawer} task. \textbf{Bottom: } selected RGB images from video clips of \texttt{pick\_up\_cup} task.}
    \label{fig:realworld_examples}
\end{figure}

\subsubsection{Real-world Experiments}

We further evaluate our method in real-world settings using a robotic manipulator performing simple tabletop manipulation tasks. Our goal is to assess the portability of our approach by capturing multi-view observations using only a handheld iPhone. Specifically, we record a video while walking around the scene, which allows us to obtain dense multi-view observations without requiring a calibrated multi-camera setup. Given strong simulation results on decision-making alongside successful real-world decomposition and novel view synthesis here, future work can explore decision-making applications on this real-world data.

\paragraph{Dataset overview.} We construct two real-world robotic manipulation scenes, \texttt{antacid\_in\_drawer} and \texttt{pick\_up\_cup} (Fig.~\ref{fig:realworld_examples}). The dataset consists of six and three video clips for the two tasks, respectively. In each clip, the robot arm is held at a fixed pose corresponding to a particular stage of the manipulation process. To increase scene diversity between each video clip, we vary both the robot arm configuration and the object arrangement to mimic snapshots of different manipulation states. This design allows the learned representation to capture the structure of the task and generalize to unseen configurations and intermediate states.

For each video clip, we split the captured trajectory into training and evaluation sets by selecting every fourth frame as a held-out evaluation view, while the remaining frames are used for training. The time-step horizons of the clips are summarized in Table~\ref{tab:video_horizons}.

\paragraph{3D reconstruction details.}
Since depth observations are unavailable in this real-world dataset, we train and evaluate our method only under the multi-view setting. We adopt the same multi-view training protocol used in RLBench, where two context views and four target views are sampled using the same view sampler. The initial context view gap is sampled from the range $[10, 100]$, and gradually increases to $[50, 200]$ as training progresses. Implementation details of the iPhone-captured real-world dataset generation pipeline are provided in Sec.~\ref{subsec:apndx_recipe}.

\subsection{Baselines}
\label{subsec:apndx_baselines}
In this section, we provide extended details on the baselines used in this work.

\subsubsection{GNFactor\texorpdfstring{~\cite{ze2023gnfactor}}{}.}
GNFactor learns 3D semantic feature volumes from single RGB-D observations using Generalizable Neural Feature Fields (GNF). These volumes, which are distilled from pre-trained 2D foundation models via multi-view rendering and Stable Diffusion features~\cite{rombach2022highresolutionimagesynthesislatent}---are processed by a Perceiver Transformer~\cite{jaegle2021perceivergeneralperceptioniterative} conditioned on language to predict multi-task actions via behavior cloning.

\subsubsection{ManiGaussian\texorpdfstring{~\cite{lu2024manigaussian}}{}.}
ManiGaussian extends Gaussian splatting with temporal dynamics and per-primitive semantic features, assuming rigid object motion. Given an observation and candidate action, its world model comprises: (i) a representation network for visual features, (ii) Gaussian parameter regression, (iii) deformation prediction for temporal evolution, and (iv) rendering for supervision. A PerceiverIO~\cite{jaegle2021perceivergeneralperceptioniterative} predicts discretized actions from the Gaussian representation and language, trained via weighted reconstruction, feature alignment, action prediction, and future-frame consistency losses.

\subsubsection{VAE-GS}
\label{subsec:vaegs}
We additionally consider a baseline, termed VAE-GS, that uses the same 3D geometry-aware encoder and Gaussian decoder as introduced in Sec.~\ref{subsec:apndx_method}, but replaces the object-centric latent space with a dense patch-level one. Specifically, the input images are first patchified and processed by the same CNN backbone. Similar to the object-centric formulation, this baseline retains a compact latent bottleneck that can be used as the visual representation for downstream imitation learning; however, this bottleneck is organized as patch-level latent features rather than object-centric particles. The resulting latent features are then upsampled back to the original patch resolution and decoded into pixel-aligned Gaussians. Accordingly, components that depend on an object-centric latent decomposition—such as attention conditioning in the encoder and particle-aligned Gaussian transformations in the decoder—are removed.

To ensure a fair comparison, we use the same patch size as in our method and train VAE-GS with an ELBO objective analogous to Sec.~\ref{sec:method_elbo}:
\begin{equation*}
\mathcal{L}_{\text{ELBO}}
=
\mathcal{L}_{\mathrm{rec}}
+
\beta_{\mathrm{KL}}\,\mathcal{L}_{\mathrm{KL}}.
\end{equation*}
Here, $\mathcal{L}_{\mathrm{rec}}$ uses the same multi-view reconstruction objective as in Sec.~\ref{sec:method_elbo}. The key difference lies in the KL term: instead of regularizing an object-centric set of latent particles with masked attribute-wise priors, VAE-GS regularizes a dense patch-level latent tensor with a standard Gaussian prior. Concretely, given multi-view observations $\mathcal{O}$, the encoder predicts a diagonal Gaussian posterior
\[
q_{\phi}(\mathcal{Z}_{\mathrm{noc}}\mid\mathcal{O})
=
\mathcal{N}\!\left(
\mathcal{Z}_{\mathrm{noc}};
\boldsymbol{\mu}_{\phi}(\mathcal{O}),
\operatorname{diag}\!\big(\boldsymbol{\sigma}_{\phi}^{2}(\mathcal{O})\big)
\right),
\]
where $\mathcal{Z}_{\mathrm{noc}}\in\mathbb{R}^{N_p\times d}$ denotes the non-object-centric patch latent representation, with $N_p$ the number of patches and $d$ the latent dimension. The decoder maps this latent code to pixel-aligned Gaussian parameters,
\[
\mathcal{G} = D_{\theta}(\mathcal{Z}_{\mathrm{noc}}),
\]
which are then rendered into context and target views.

The KL term is therefore defined as
\[
\mathcal{L}_{\mathrm{KL}}
=
D_{\mathrm{KL}}\!\left(
q_{\phi}(\mathcal{Z}_{\mathrm{noc}}\mid\mathcal{O})
\,\Vert\,
p(\mathcal{Z}_{\mathrm{noc}})
\right),
\quad
p(\mathcal{Z}_{\mathrm{noc}})
=
\mathcal{N}(\mathbf{0}, \mathbf{I}),
\]
and, since both posterior and prior are diagonal Gaussians, it is computed in closed form as
\[
D_{\mathrm{KL}}\!\left(
q_{\phi}(\mathcal{Z}_{\mathrm{noc}}\mid\mathcal{O})
\,\Vert\,
p(\mathcal{Z}_{\mathrm{noc}})
\right)
=
\frac{1}{2}
\sum_j
\left(
\mu_j^2 + \sigma_j^2 - \log \sigma_j^2 - 1
\right).
\]
This baseline isolates the effect of replacing the object-centric latent space with a compact non-object-centric patch bottleneck while keeping the same 3D geometric conditioning and rendering framework.

\subsection{Hyperparameters and Training Details}
\label{subsec:apndx_hp}

\paragraph{Training.}
We implement our method in PyTorch~\cite{paszke2017pytorch} and optimize all models using Adam~\cite{adam_14} with $\beta_1 = 0.9$, $\beta_2 = 0.999$, and $\epsilon = 10^{-6}$. Unless otherwise specified, we use a constant learning rate of $8 \times 10^{-5}$ and a batch size of 8, and train for 50{,}000 gradient steps.  Training is performed on a single A6000 GPU and takes approximately 30 hours for RLBench dataset, and 2 hours for the real-world dataset.

\paragraph{Data preprocessing and augmentation.}
All input images are center-cropped and resized to $128 \times 128$. During training, we apply random horizontal flipping as data augmentation. We normalize camera intrinsics, and convert camera extrinsics to the camera-to-world (\texttt{c2w}) convention.

\paragraph{Hyperparameters.}
We provide the full set of hyperparameters in Table~\ref{tab:hyperparams}, with separate configurations reported for RLBench and real-world experiments.

\begin{table}[t]
\centering
\scriptsize
\renewcommand{\arraystretch}{1.05}
\setlength{\tabcolsep}{6pt}
\begin{tabular}{lcc}
\toprule
\textbf{Hyperparameter} & \textbf{RLBench} & \textbf{Real-world} \\
\midrule
$L$ (\# Particles)          & 20                          & 40 \\
$M$ (\# KP Proposals)       & 256                         & 256 \\
Reconstruction Loss         & MSE                         & LPIPS \\
$\beta_{\mathrm{KL}}$       & 0.02                        & 0.1 \\
$\beta_{\mathrm{KP}}$       & 0.02                        & 0.01 \\
$\lambda_{D}$               & 0.001                       & - \\
KP Proposal Patch Size      & 8                           & 8 \\
Glimpse Ratio               & 0.125                       & 0.25 \\
$d_{\mathrm{fg}}$           & 16                          & 16 \\
$d_{\mathrm{bg}}$           & 5                           & 16 \\
$d_{\mathrm{proj}}$         & 512                         & 512 \\
FG CNN Ch. Mult.            & $[1,4,8]$                   & $[1,1,4,8]$ \\
BG CNN Ch. Mult.            & $[1,1,1,4,8]$               & $[1,1,1,4,8]$ \\
Particle Interaction Transformer Layers & 6             & 6 \\
Particle Interaction Transformer Heads  & 8             & 8 \\
\midrule
SH Degree                   & 1                           & 4 \\
$d_{\max}$                  & 5.0                   & 10.0 \\
$d_{\min}$                  & 0.5                   & 0.01 \\
$s_{\max}$                  & 15.0                   & 15.0 \\
$s_{\min}$                  & 0.5                 & 0.5 \\
\bottomrule
\end{tabular}
\caption{\textbf{Hyperparameters across datasets.} The top section lists model hyperparameters, while the bottom section lists renderer hyperparameters. We set the base CNN channel count to 32 for all experiments.}
\label{tab:hyperparams}
\end{table}

\begin{table*}[t]
\centering
\scriptsize
\renewcommand{\arraystretch}{1.1}
\setlength{\tabcolsep}{6pt}
\begin{tabular}{l l c c}
\toprule
\textbf{Attribute} & \textbf{Distribution} & \textbf{RLBench} & \textbf{Real-world} \\
\midrule
Position Offset $z_o$
& Normal, $\mathcal{N}(\mu, \sigma^2)$
& $\mu = 0, \ \sigma = 0.2$
& $\mu = 0, \ \sigma = 0.1$ \\

Scale $z_s$
& Normal, $\mathcal{N}(\mu, \sigma^2)$
& \begin{tabular}[c]{@{}c@{}}
$\mu = \mathrm{Sigmoid}^{-1}(0.25)$ \\
$\sigma = 0.3$
\end{tabular}
& \begin{tabular}[c]{@{}c@{}}
$\mu = \mathrm{Sigmoid}^{-1}(0.125)$ \\
$\sigma = 0.15$
\end{tabular} \\

Depth $z_d$
& Normal, $\mathcal{N}(\mu, \sigma^2)$
& $\mu = 0, \ \sigma = 1$
& $\mu = 0, \ \sigma = 1$ \\

Depth Extent $z_e$
& Normal, $\mathcal{N}(\mu, \sigma^2)$
& \begin{tabular}[c]{@{}c@{}}
$\mu = \mathrm{Sigmoid}^{-1}(0.3)$ \\
$\sigma = 0.3$
\end{tabular}
& \begin{tabular}[c]{@{}c@{}}
$\mu = \mathrm{Sigmoid}^{-1}(0.3)$ \\
$\sigma = 0.3$
\end{tabular} \\

Transparency $z_t$
& Beta, $\mathrm{Beta}(a, b)$
& $a = 0.01, \ b = 0.01$
& $a = 0.01, \ b = 0.01$ \\

Appearance Features $z_f, z_{\mathrm{bg}}$
& Normal, $\mathcal{N}(\mu, \sigma^2)$
& $\mu = 0, \ \sigma = 1$
& $\mu = 0, \ \sigma = 1$ \\
\bottomrule
\end{tabular}
\caption{\textbf{Prior distribution parameters across datasets.} We use dataset-specific prior settings for the latent attributes in our object-centric representation.}
\label{tab:hyperparams_prior}
\end{table*}

\subsection{Extended Experiment Results}

\subsubsection{Representation compactness and efficiency}
\label{subsec:comp-effic}
\paragraph{Latent space compactness}
We present an explicit compactness analysis against ManiGaussian in the \emph{single-view RLBench RGB-D setting} used for policy-facing evaluation. Under the same output Gaussian field used for rendering ($128^2$ predicted Gaussians, 23 parameters per Gaussian), ParticleSplat stores the scene in only 20 latent particles of 23 dimensions each, yielding a compression ratio of $(128^2\!\cdot\!23)/(20\!\cdot\!23)=819.2\times$. In contrast, ManiGaussian uses a dense pre-Gaussian voxel feature grid: with voxel shape $100$, patch size $5$, and 128-D features, the effective state is $(100/5)^3\!\times\!128=20^3\!\times\!128$, giving $(128^2\!\cdot\!23)/(20^3\!\cdot\!128)=0.368\times$.
\paragraph{Runtime analysis}
Although Gaussians are decoded per view and stitched into a global set, the total effective number of gaussians is limited by the finite particle budget, opacity gating, and scene-level KL. In single-view RGB-D, ParticleSplat uses 251.1 MiB peak CUDA memory, 30.91 ms model-only latency, and 32.35 samples/s; with two views, it uses 347.3 MiB, 33.65 ms, and 29.72 samples/s. This shows practical policy learning applications under the real-time setting and efficient scaling with input views.

\subsubsection{EC-Diffuser with language}
\label{subsec:ecdiff-lang}
In its original formulation, EC-Diffuser encodes each observation into a set of entity tokens, appends the action as an additional token, and trains a conditional diffusion model to denoise future entity states and actions conditioned on the current state and a goal state. In RLBench, however, the task is specified by a natural language instruction rather than a goal observation. We therefore replace goal conditioning with \emph{language conditioning}.

In our implementation, we use the self-attention variant of the language-conditioned denoiser. Specifically, the frozen visual encoder first maps each observation to a compact set of latent tokens, which serve as the state representation for the policy. The language instruction is encoded using a pre-trained CLIP text encoder, projected into the same transformer dimension, and treated as a set of additional tokens. That is, we treat the language tokens as \emph{additional particles} and concatenate them with the action token and visual latent tokens before passing them to the transformer.

The denoiser then jointly processes action, visual, and language tokens through self-attention over the planning horizon, while the diffusion timestep is injected through adaptive layer normalization. In this way, task semantics from language can directly interact with the visual latent representation and action tokens within a shared token space, without requiring a separate cross-attention module. The model is trained with the standard diffusion denoising objective used in EC-Diffuser.

During training, the visual encoder is frozen after representation learning, and only the diffusion policy is optimized on top of the extracted latent features. At inference time, the model predicts a sequence of future actions conditioned on the current encoded observation and language instruction, and only the first action is executed before replanning, following the same receding-horizon control scheme as EC-Diffuser.

\begin{figure}
    \centering
    \includegraphics[width=\linewidth]{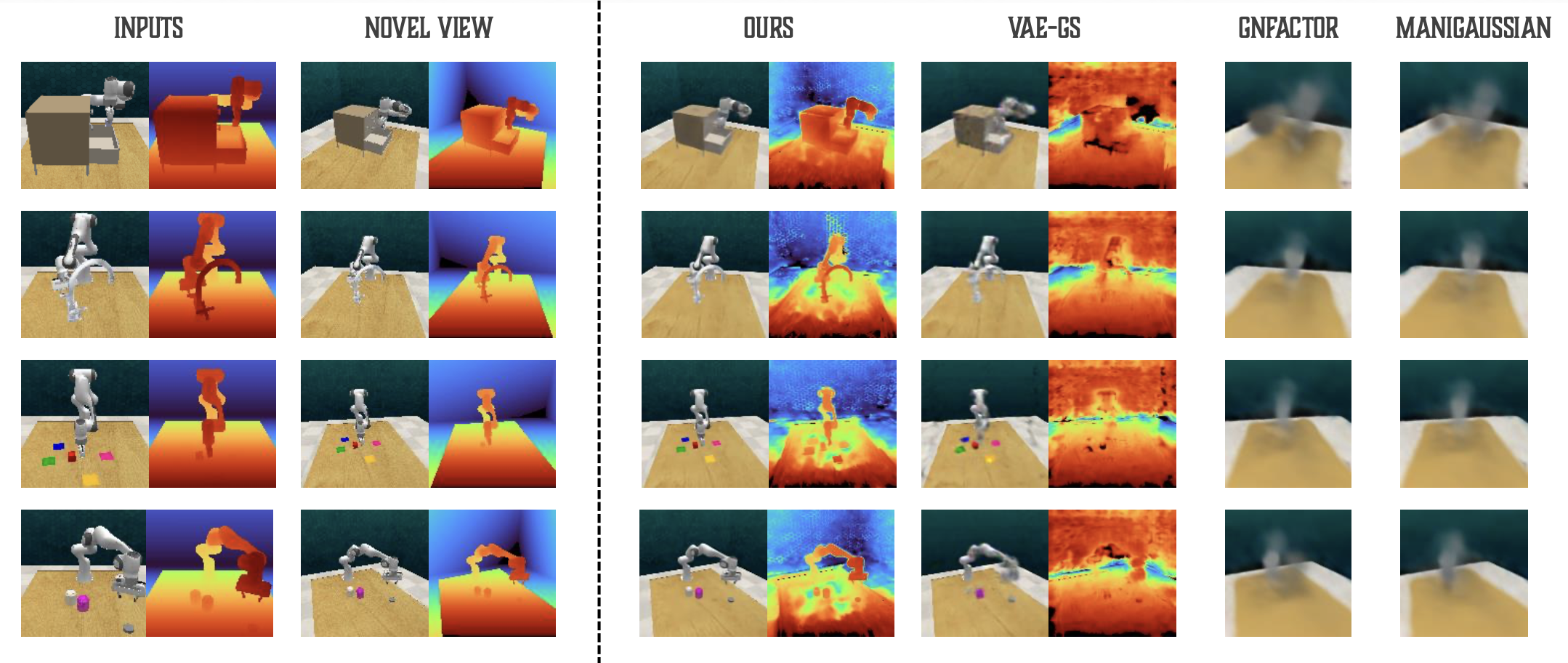}
    \caption{\textbf{Extended qualitative Novel View Synthesis Results from RLBench scenes.} We provide additional visualizations for novel view renderings produced by our learned representation as well as other baselines on RLBench scenes.}
    \label{fig:rlbench_nvs_ext}
\end{figure}

\begin{figure}
    \centering
    \includegraphics[width=\linewidth]{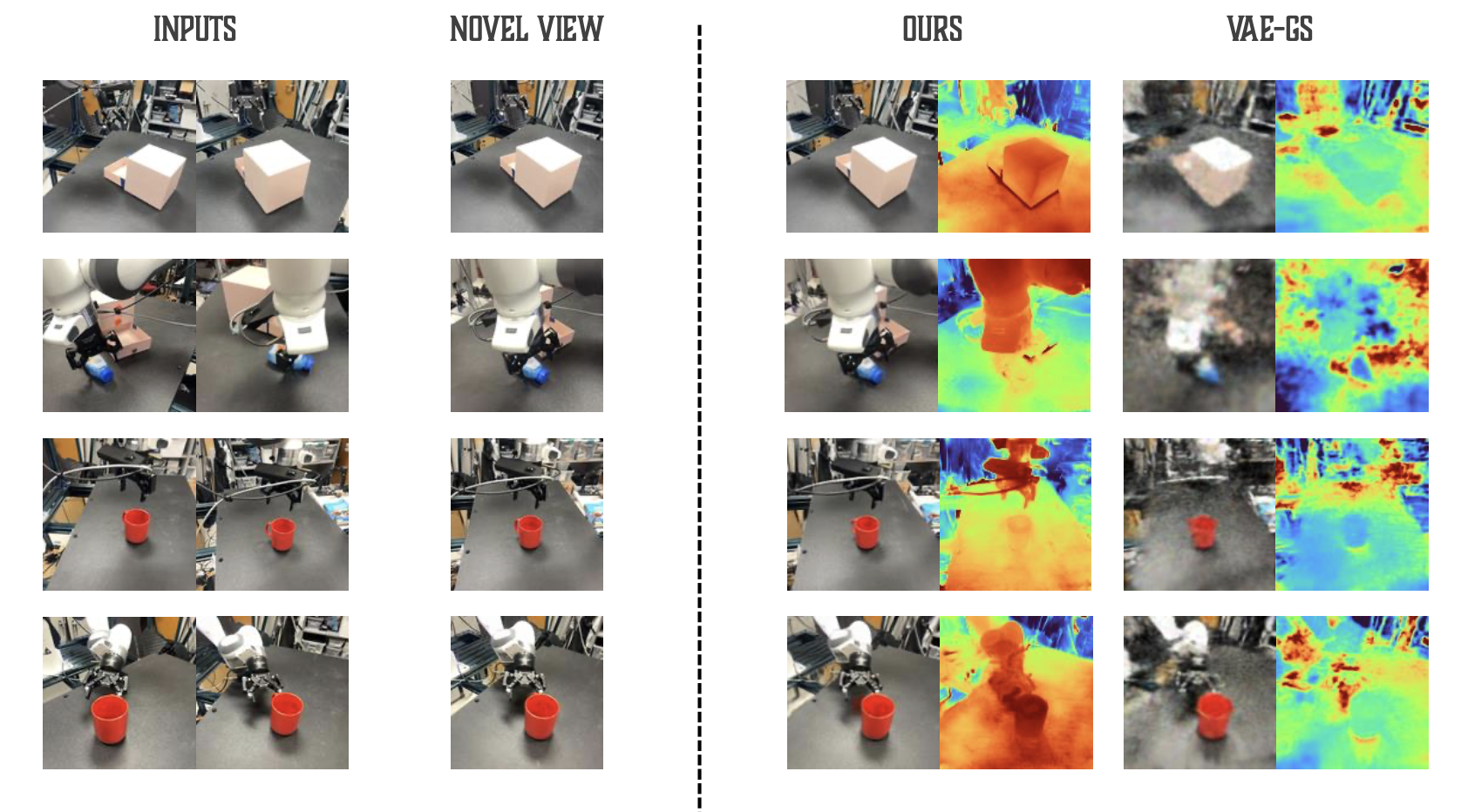}
    \caption{\textbf{Extended qualitative Novel View Synthesis Results from Real-world scenes.} We provide additional visualizations for novel view renderings produced by our learned representation as well as other learned baselines on real-world scenes.}
    \label{fig:realworld_nvs_ext}
\end{figure}

\subsubsection{Novel View Synthesis -- Extended Results}
\label{subsec:nvs-extended}
\begin{figure}
    \centering
    \includegraphics[width=1\linewidth]{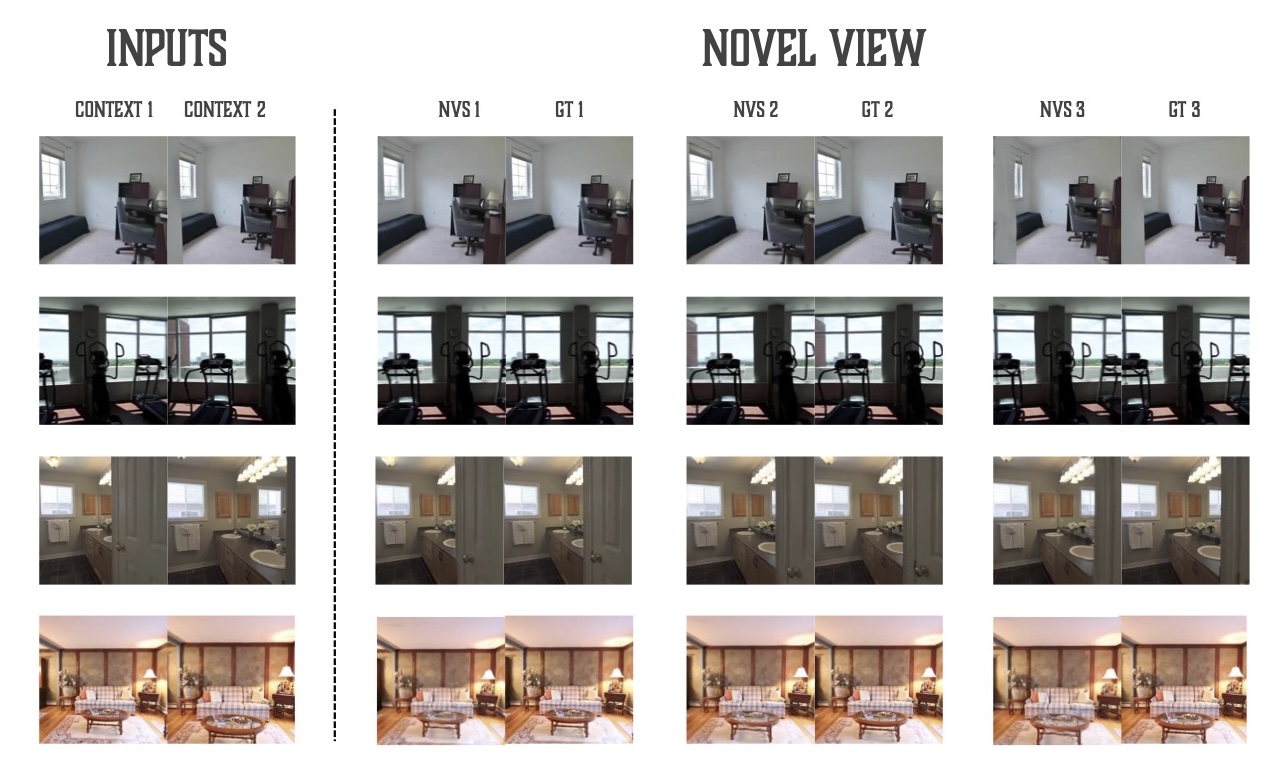}
    \caption{\textbf{Additional Novel View Synthesis Results from RE10k subset.} We provide additional visualizations for novel view renderings produced by our learned representation on the subset of RE10k dataset.}
    \label{fig:re10k_nvs_ext}
\end{figure}

This section extends the novel view synthesis results of Sec.~\ref{subsec:exp_novelv}. We provide additional qualitative comparisons on RLBench and real-world scenes (Figs.~\ref{fig:rlbench_nvs_ext} and~\ref{fig:realworld_nvs_ext}), an ablation of the 3D-aware priors in the encoder under the single-view and multi-view settings (Table~\ref{tab:ablation_enc}), and results on a subset of RE10K at $256\times256$ resolution (Fig.~\ref{fig:re10k_nvs_ext}).

\paragraph{Additional qualitative results.}
We provide extended qualitative results for novel view synthesis on RLBench scenes in Fig.~\ref{fig:rlbench_nvs_ext} and real-world scenes in Fig.~\ref{fig:realworld_nvs_ext}. Specifically, we include visualizations on RLBench under the single-view RGB-D setting to compare our method against competing baselines, as well as results on real-world scenes under the multi-view RGB setting. As shown in the figures, our method consistently produces renderings with high visual fidelity, preserving fine scene details and maintaining geometric consistency across viewpoints. Notably, these results highlight the benefit of our object-centric representation: by explicitly structuring the scene around coherent object-level entities, the learned representation captures both appearance and geometry in a way that supports more faithful view synthesis and more interpretable scene understanding. This is further reflected in the predicted depth, which remains visually interpretable and physically plausible, exhibiting coherent object boundaries and reasonable spatial layout. Together, these qualitative results suggest that the object-centric design of our representation contributes to strong generalization across both simulated and real-world environments while preserving both appearance quality and 3D structural consistency.

\paragraph{Ablations analysis - encoder.}
We ablate the \textbf{3D-aware priors in our encoder} through novel view synthesis, since this evaluation directly measures how well the learned representation captures geometrically consistent scene structure. Rather than assessing downstream control, this experiment isolates the contribution of encoder-side 3D cues to multi-view consistency by comparing the full model against variants without pose input and without depth input. Table~\ref{tab:ablation_enc} reports the resulting novel view synthesis performance under both single-view and multi-view settings.

Here, \textbf{w/o pose} removes all camera-pose conditioning from the encoder, including both the FiLM-based attention conditioning and the convolutional conditioning with pose-derived geometric features. In contrast, \textbf{w/o depth} removes depth from the convolutional conditioning pathway and also removes depth supervision from the reconstruction loss, so the model is trained purely from RGB reconstruction. All other components are kept unchanged.

Several trends are clear from Table~\ref{tab:ablation_enc}. First, pose conditioning is consistently effective across both settings: removing it degrades performance in both single-view and multi-view reconstruction, confirming that explicit camera geometry provides a strong inductive bias for learning view-consistent scene representations. This result supports our design choice of injecting pose information into both convolutional features and particle interactions.

Second, depth supervision is essential in the single-view setting. When only one context view is available, removing depth causes a large drop in reconstruction quality. This is expected, since monocular reconstruction is fundamentally ambiguous, and explicit depth supervision and conditioning provide a critical signal for recovering 3D structure from a single observation. In contrast, depth is much less critical in the multi-view setting. With two context views, the performance gap between the full model and the variant without depth is comparatively small. Intuitively, under multi-view supervision the model can infer much of the scene geometry from cross-view consistency alone. In particular, the object-centric latent formulation encourages particles to align across views and explain a shared underlying 3D scene, which provides a form of self-supervised geometric alignment even without explicit depth supervision. As a result, the model can still learn meaningful depth structure from RGB-only multi-view reconstruction, although explicit depth remains beneficial.

\begin{table*}[t]
    \centering
    \scriptsize
    \begin{tabular}{l|l|c|ccc}
        \toprule
        Dataset & Method & Input & PSNR $\uparrow$ & SSIM $\uparrow$ & LPIPS $\downarrow$ \\
        \midrule
        \multirow{6}{*}{RLBench}
        & ParticleSplat (Ours) w/o depth & 1$\times$RGB & 19.63 & 0.4248 & 0.6836 \\
        & ParticleSplat (Ours) w/o pose  & 1$\times$RGB-D & 23.91 & 0.6815 & 0.4012 \\
        & \textbf{ParticleSplat (Ours)}  & 1$\times$RGB-D & \textbf{25.35} & \textbf{0.7293} & \textbf{0.3563} \\
        \cmidrule(l){2-6}
        & ParticleSplat (Ours) w/o depth & 2$\times$RGB & 28.38 & 0.7862 & 0.2894 \\
        & ParticleSplat (Ours) w/o pose  & 2$\times$RGB-D & 27.62 & 0.7798 & 0.3016 \\
        & \textbf{ParticleSplat (Ours)}  & 2$\times$RGB-D & \textbf{28.77} & \textbf{0.8144} & \textbf{0.2565} \\
        \bottomrule
    \end{tabular}
    \caption{\textbf{Ablation on RLBench novel view synthesis.} We compare the full model with variants without camera pose input and without depth input under single-view and multi-view settings. We report PSNR$\uparrow$, SSIM$\uparrow$, and LPIPS$\downarrow$.}
    \label{tab:ablation_enc}
\end{table*}

\paragraph{Additional RE10K subset results.}
Although ParticleSplat is designed primarily as a self-supervised object-centric representation learning method for downstream decision making, we additionally evaluate it on the RE10K subset provided by MVSplat~\cite{chen2024mvsplat} to assess whether the learned particle-based representation generalizes beyond robot-tabletop scenes. This setting contains more diverse and visually complex in-the-wild scenes, and we conduct these experiments at $256\times256$ resolution to examine whether our compact particle bottleneck can preserve higher-frequency appearance and view-consistent geometry. As shown in Fig.~\ref{fig:re10k_nvs_ext}, ParticleSplat produces coherent novel-view renderings on held-out test scenes and achieves 24.21 PSNR, 0.816 SSIM, and 0.177 LPIPS. While under-performing RE10K-specialized SOTA due to the smaller training/evaluation subset, these results demonstrate that the proposed representation remains effective beyond the manipulation-focused distributions used in the main experiments.

\subsubsection{Self-supervised Object-centric 3D scene Decomposition - Extended results}
\begin{figure}
    \centering
    \includegraphics[width=\linewidth]{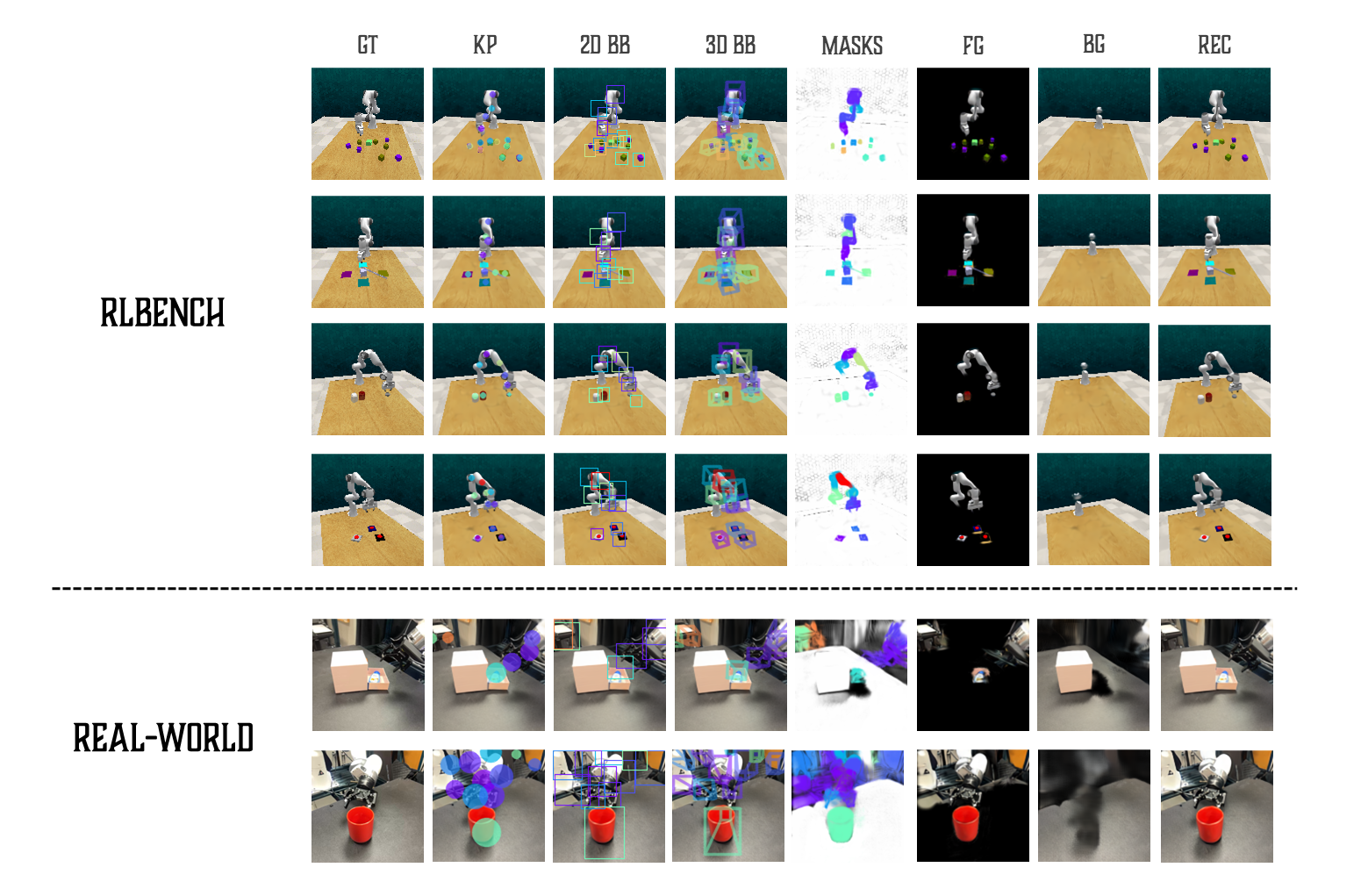}
    \caption{\textbf{Additional scene decompositions with ParticleSplat.} Additional self-supervised object-centric
decomposition on both synthetic RLBench data and real-world data.}
    \label{fig:latent_decomp_ext}
\end{figure}
We present additional object-centric decompositions of 3D scenes in Fig.~\ref{fig:latent_decomp_ext}. Across both the synthetic RLBench dataset and the real-world dataset, our method yields clean 3D entity decompositions on unseen test scenes, highlighting its strong generalization capability. In the cluttered real-world setting, we apply non-maximum suppression as a post-processing step to remove redundant particles with overlapping bounding boxes and improve visual clarity. Additional videos of the novel view rendering of the decompositions can be found in the supplementary.

\subsubsection{Imitation Learning - Extended results}
\label{subsec:il-extended}
This section extends the imitation-learning results of Sec.~\ref{subsec:exp_imit}. We report per-task success rates on RLBench in the single-view setting (Table~\ref{tab:manipulation_results_full}), results in the multi-view setting together with an ablation of the representation under both settings (Table~\ref{tab:ablation_repr}), and the per-task MimicGen comparison against the state-of-the-art methods of that benchmark (Table~\ref{tab:mimicgen_full}), together with its experimental setup and result analysis.

\begin{table*}[!t]
\scriptsize
\centering
\begin{tabular*}{\textwidth}{@{\extracolsep{\fill}} l cccccc @{}}
\toprule
\textbf{Method} / \textit{Task}
& \begin{tabular}[c]{@{}c@{}}\textit{push}\\\textit{buttons}\end{tabular}
& \begin{tabular}[c]{@{}c@{}}\textit{meat off}\\\textit{grill}\end{tabular}
& \begin{tabular}[c]{@{}c@{}}\textit{slide}\\\textit{block}\end{tabular}
& \begin{tabular}[c]{@{}c@{}}\textit{drag}\\\textit{stick}\end{tabular}
& \begin{tabular}[c]{@{}c@{}}\textit{sweep to}\\\textit{dustpan}\end{tabular} \\
\midrule
GNFactor~\cite{ze2023gnfactor} & 18.7{\scriptsize$\pm$10.0} & 57.3{\scriptsize$\pm$18.9} & 20.0{\scriptsize$\pm$15.0} & 37.3{\scriptsize$\pm$13.2} & 28.0{\scriptsize$\pm$15.0}\\
ManiGaussian~\cite{lu2024manigaussian} & 20.4{\scriptsize$\pm$12.2} & 60.2{\scriptsize$\pm$18.2} & 24.3{\scriptsize$\pm$12.8} & 92.3{\scriptsize$\pm$11.4} & 64.5{\scriptsize$\pm$13.4}\\
VAE-GS & 22.3{\scriptsize$\pm$11.4} & 58.8{\scriptsize$\pm$7.9} & 25.7{\scriptsize$\pm$10.9} & 90.6{\scriptsize$\pm$7.8} & 46.9{\scriptsize$\pm$8.1}\\
DLP~\cite{daniel2026lpwm} & \underline{36.7}{\scriptsize$\pm$8.9} & \underline{65.4}{\scriptsize$\pm$4.1} & \underline{54.6}{\scriptsize$\pm$5.2} & \underline{93.1}{\scriptsize$\pm$5.5} & \underline{73.5}{\scriptsize$\pm$7.3}\\
\textbf{ParticleSplat (Ours)} & \textbf{57.2}{\scriptsize$\pm$7.1} & \textbf{76.0}{\scriptsize$\pm$8.4} & \textbf{92.3}{\scriptsize$\pm$2.0} & \textbf{96.7}{\scriptsize$\pm$2.5} & \textbf{90.6}{\scriptsize$\pm$6.9}\\
\midrule
\textbf{Method} / \textit{Task} 
& \begin{tabular}[c]{@{}c@{}}\textit{turn}\\\textit{tap}\end{tabular}
& \begin{tabular}[c]{@{}c@{}}\textit{open}\\\textit{drawer}\end{tabular}
& \begin{tabular}[c]{@{}c@{}}\textit{put in}\\\textit{drawer}\end{tabular}
& \begin{tabular}[c]{@{}c@{}}\textit{stack}\\\textit{blocks}\end{tabular}
& \begin{tabular}[c]{@{}c@{}}\textit{close}\\\textit{jar}\end{tabular} 
& \multicolumn{1}{c}{\textbf{Overall}} \\
\midrule
GNFactor~\cite{ze2023gnfactor} & 50.7{\scriptsize$\pm$8.2} & 76.0{\scriptsize$\pm$5.7} & 0.0{\scriptsize$\pm$0.0} & 4.0{\scriptsize$\pm$3.3} & 25.3{\scriptsize$\pm$6.8} & 31.7 \\
ManiGaussian~\cite{lu2024manigaussian} & \underline{56.2}{\scriptsize$\pm$6.8} & \underline{76.3}{\scriptsize$\pm$6.2} & 16.6{\scriptsize$\pm$3.2} & 12.4{\scriptsize$\pm$3.2} & 28.4{\scriptsize$\pm$5.4} & 45.2 \\
VAE-GS (EC-Diff.) & 52.5{\scriptsize$\pm$6.3} & 73.1{\scriptsize$\pm$5.4} & 12.1{\scriptsize$\pm$3.0} & 8.6{\scriptsize$\pm$3.0} & \underline{29.6}{\scriptsize$\pm$5.0} & 42.0 \\
DLP~\cite{daniel2026lpwm} (EC-Diff.) & 36.2{\scriptsize$\pm$5.8} & 37.8{\scriptsize$\pm$9.6} & \underline{30.8}{\scriptsize$\pm$4.7} & \underline{11.4}{\scriptsize$\pm$4.6} & 20.7{\scriptsize$\pm$4.9} & \underline{46.6} \\
\textbf{ParticleSplat (Ours)} & \textbf{83.8}{\scriptsize$\pm$4.6} & \textbf{88.4}{\scriptsize$\pm$4.3} & \textbf{78.4}{\scriptsize$\pm$5.8} & \textbf{19.4}{\scriptsize$\pm$2.9} & \textbf{54.1}{\scriptsize$\pm$4.7} & \textbf{73.0} \\
\bottomrule
\end{tabular*}
\caption{\textbf{Per-task imitation learning performance with different representations (single-view).} Success rates (\%) on RLBench manipulation tasks using policies conditioned on different learned representations. Results are mean $\pm$ standard deviation over three seeds (25 episodes per task per seed); Table~\ref{tab:manipulation_results} reports the per-category averages}
\label{tab:manipulation_results_full}
\vskip -0.12in
\end{table*}

\begin{table*}[!t]
\scriptsize
\centering
\begin{tabular*}{\textwidth}{@{\extracolsep{\fill}} l ccccccc @{}}
\toprule
\textbf{Method} / \textit{Task}
& \textit{Stack}
& \begin{tabular}[c]{@{}c@{}}\textit{Stack}\\\textit{Three}\end{tabular}
& \textit{Square}
& \textit{Threading}
& \textit{Coffee}
& \begin{tabular}[c]{@{}c@{}}\textit{Three}\\\textit{Pc. Asmbl.}\end{tabular} & \\
\midrule
DLP~\cite{daniel2026lpwm} (multi-view) & 78.0{\scriptsize$\pm$2.8} & 14.7{\scriptsize$\pm$6.2} & 45.3{\scriptsize$\pm$10.5} & \underline{45.3}{\scriptsize$\pm$6.8} & \underline{82.0}{\scriptsize$\pm$2.8} & 29.3{\scriptsize$\pm$7.7} & \\
3D-DLP~\cite{zhang20263ddlp} & \textbf{94.6}{\scriptsize$\pm$0.9} & \textbf{70.0}{\scriptsize$\pm$1.6} & \underline{51.3}{\scriptsize$\pm$0.9} & 36.0{\scriptsize$\pm$1.6} & 36.0{\scriptsize$\pm$1.6} & \underline{38.0}{\scriptsize$\pm$4.0} & \\
EquiDiff~\cite{wang2024equivariant} (voxel only) & 82.0{\scriptsize$\pm$0.0} & 12.7{\scriptsize$\pm$5.7} & 50.0{\scriptsize$\pm$2.8} & 42.0{\scriptsize$\pm$0.0} & 70.7{\scriptsize$\pm$3.4} & 31.3{\scriptsize$\pm$3.7} & \\
\textbf{ParticleSplat (Ours)} & \underline{92.7}{\scriptsize$\pm$2.5} & \underline{68.0}{\scriptsize$\pm$4.3} & \textbf{58.7}{\scriptsize$\pm$5.0} & \textbf{54.0}{\scriptsize$\pm$4.3} & \textbf{89.7}{\scriptsize$\pm$3.4} & \textbf{46.7}{\scriptsize$\pm$4.1} & \\
\midrule
\textbf{Method} / \textit{Task}
& \begin{tabular}[c]{@{}c@{}}\textit{Hammer}\\\textit{Cl.}\end{tabular}
& \begin{tabular}[c]{@{}c@{}}\textit{Mug}\\\textit{Cl.}\end{tabular}
& \textit{Kitchen}
& \begin{tabular}[c]{@{}c@{}}\textit{Nut}\\\textit{Asmbl.}\end{tabular}
& \begin{tabular}[c]{@{}c@{}}\textit{Pick}\\\textit{Place}\end{tabular}
& \begin{tabular}[c]{@{}c@{}}\textit{Coffee}\\\textit{Prep.}\end{tabular}
& \multicolumn{1}{c}{\textbf{Overall}} \\
\midrule
DLP~\cite{daniel2026lpwm} (multi-view) & 66.7{\scriptsize$\pm$2.5} & 34.7{\scriptsize$\pm$7.5} & 0.0{\scriptsize$\pm$0.0} & 8.7{\scriptsize$\pm$2.5} & \underline{4.7}{\scriptsize$\pm$2.5} & 0.0{\scriptsize$\pm$0.0} & 34.1 \\
3D-DLP~\cite{zhang20263ddlp} & \textbf{94.6}{\scriptsize$\pm$0.9} & \textbf{64.0}{\scriptsize$\pm$4.3} & \underline{86.7}{\scriptsize$\pm$3.4} & 6.0{\scriptsize$\pm$1.6} & 0.0{\scriptsize$\pm$0.0} & 0.0{\scriptsize$\pm$0.0} & \underline{48.1} \\
EquiDiff~\cite{wang2024equivariant} (voxel only) & \underline{92.6}{\scriptsize$\pm$3.3} & 34.0{\scriptsize$\pm$2.8} & \textbf{95.0}{\scriptsize$\pm$2.5} & \underline{10.7}{\scriptsize$\pm$2.1} & \textbf{12.7}{\scriptsize$\pm$2.2} & \textbf{34.0}{\scriptsize$\pm$4.9} & 47.3 \\
\textbf{ParticleSplat (Ours)} & 90.0{\scriptsize$\pm$2.8} & \underline{62.0}{\scriptsize$\pm$4.9} & 76.0{\scriptsize$\pm$5.7} & \textbf{16.0}{\scriptsize$\pm$3.3} & \textbf{12.7}{\scriptsize$\pm$3.4} & \underline{6.7}{\scriptsize$\pm$2.5} & \textbf{56.1} \\
\bottomrule
\end{tabular*}
\caption{\textbf{Per-task imitation learning performance on MimicGen (multi-view).} Success rates (\%) on 12 single-arm MimicGen~\cite{mandlekar2023mimicgen} tasks; representation-learning methods (DLP, 3D-DLP, ParticleSplat) share the same EC-Diffuser policy, whereas EquiDiff is a voxel-based diffusion policy; mean $\pm$ standard deviation over three seeds (50 rollouts per task per seed), 200 demonstrations per task. For a thorough comparison against the state-of-the-art methods on this benchmark, we include 3D-DLP~\cite{zhang20263ddlp} and the voxel-based EquiDiff~\cite{wang2024equivariant} policy; their results and the multi-view DLP results are reproduced from~\cite{zhang20263ddlp}. Tasks are ordered by the benchmark's official groups: Basic (\textit{Stack}, \textit{Stack Three}), Contact-Rich (\textit{Square} to \textit{Mug Cleanup}), and Long-Horizon (\textit{Kitchen} to \textit{Coffee Preparation}); Table~\ref{tab:mimicgen} reports the per-group averages.}
\label{tab:mimicgen_full}
\vskip -0.12in
\end{table*}

\newpage
\paragraph{Multi-view imitation learning.} Unlike prior works using monocular RGB-D, we leverage multi-view inputs at inference. Since extending other policies to multi-view is nontrivial, we compare only against VAE-GS and DLP.
Table~\ref{tab:ablation_repr} shows that ParticleSplat remains the strongest representation under the multi-view setting, achieving an overall success rate of 83.0\%, compared with 59.4\% for DLP and 52.7\% for VAE-GS. The advantage is consistent across all ten tasks, including object-centric interaction tasks such as \texttt{push\_buttons}, \texttt{meat\_off\_grill}, and \texttt{slide\_block}, geometry-sensitive tasks such as \texttt{open\_drawer} and \texttt{turn\_tap}, and tasks that require both accurate spatial reasoning and object-level understanding, such as \texttt{close\_jar}, \texttt{put\_in\_drawer}, and \texttt{stack\_blocks}. While the relative benefit of additional viewpoints varies across methods and tasks, the main observations are that additional viewpoints benefit all methods, and ParticleSplat continues to deliver the best downstream manipulation performance in the multi-view regime. This suggests that explicit 3D object-centric structure remains a strong inductive bias even when richer visual input is available.

\paragraph{Ablation analysis.}
We ablate our \textbf{3D-aware object-centric representation} through downstream decision-making performance across both single-view and multi-view settings, as this evaluation enables a direct comparison among representations that span the key design dimensions relevant to manipulation: 2D object-centric structure, non-object-centric representations trained to encode 3D information, and our combined object-centric formulation that explicitly captures 3D scene information. Table~\ref{tab:ablation_repr} isolates the effect of representation design by comparing DLP, VAE-GS, and ParticleSplat under both single-view and multi-view settings. DLP outperforms VAE-GS in both single-view (46.6\% vs.\ 42.0\%) and multi-view (59.4\% vs.\ 52.7\%), suggesting that object-centric decomposition is more beneficial for downstream manipulation than a non-object-centric patch-based latent space, even when the latter uses the same encoder and decoder family as our method and is trained to encode 3D information. In contrast, ParticleSplat substantially surpasses both baselines, reaching 73.0\% in the single-view setting and 83.0\% in the multi-view setting. This result indicates that the performance gain cannot be attributed solely to the reconstruction objective or decoder architecture. Rather, the advantage comes from combining \emph{object-centric structure} with \emph{explicit 3D scene information}. By representing the scene as object-centric 3D Gaussians, ParticleSplat produces features that are better aligned with both interaction-centric reasoning and geometric control, leading to improved policy learning in both sparse-view and multi-view regimes.

\paragraph{Additional MimicGen results.}
We additionally evaluate ParticleSplat on MimicGen~\cite{mandlekar2023mimicgen} alongside a contemporary 3D object-centric representation method (3D-DLP) and an end-to-end voxel policy (EquiDiff) for a holistic picture of where ParticleSplat's learned representation helps (Table~\ref{tab:mimicgen_full}, reproduced from~\cite{zhang20263ddlp}). 3D-DLP~\cite{zhang20263ddlp} is the method closest to ours: it also lifts DLP particles to 3D (position, scale, transparency, appearance) and feeds frozen particle tokens to the same EC-Diffuser policy, but it operates on a dense $64^3$ RGB voxel grid built from the static cameras' point clouds, whereas ParticleSplat infers particles directly from posed images and decodes them into splatted 3D Gaussians, never materialising a volumetric grid. EquiDiff (voxel-only)~\cite{wang2024equivariant} is not a representation but a strong SE(3)-equivariant diffusion policy acting on the same voxels. All methods use 200 D0 demonstrations per task, the same two static third-person cameras (no eye-in-hand), 50 rollouts per task over three seeds, and the adapted EC-Diffuser backbone for the particle methods (details in~\cite{zhang20263ddlp}). ParticleSplat receives exactly the baselines' context views. Similar to RLBench, additional rendered views serve only as reconstruction targets during representation learning. As shown in Table~\ref{tab:mimicgen_full}, ParticleSplat attains the best overall success rate (56.1\% vs.\ 48.1\% for 3D-DLP and 47.3\% for EquiDiff), wins five tasks and ties \textit{Pick Place}, with the largest margin on the contact-rich group (66.9\% vs.\ 53.3\%/53.4\%). These tasks hinge on sub-centimetre alignment of small parts whose geometric evidence a $64^3$ voxelisation (1--2\,cm per cell) largely averages away, whereas ParticleSplat retains it at image resolution and refines particle position and scale through multi-view rendering consistency; 3D-DLP stays 2--5 points ahead on stacking and cleanup, and the equivariant voxel policy leads the long-horizon \textit{Kitchen} and \textit{Coffee Preparation}.

\begin{table*}[t]
\scriptsize
\centering

\begin{tabular*}{\textwidth}{@{\extracolsep{\fill}} l l cccccc @{}}
\toprule
\textbf{Setting} & \textbf{Method}/ \textit{Task}
& \begin{tabular}[c]{@{}c@{}}\textit{close}\\\textit{jar}\end{tabular}
& \begin{tabular}[c]{@{}c@{}}\textit{open}\\\textit{drawer}\end{tabular}
& \begin{tabular}[c]{@{}c@{}}\textit{sweep to}\\\textit{dustpan}\end{tabular}
& \begin{tabular}[c]{@{}c@{}}\textit{meat off}\\\textit{grill}\end{tabular}
& \begin{tabular}[c]{@{}c@{}}\textit{turn}\\\textit{tap}\end{tabular} \\
\midrule

\multirow{3}{*}{Single-view}
& VAE-GS
& 29.6{\scriptsize$\pm$5.0}
& 73.1{\scriptsize$\pm$5.4}
& 46.9{\scriptsize$\pm$8.1}
& 58.8{\scriptsize$\pm$7.9}
& 52.5{\scriptsize$\pm$6.3} \\

& DLP~\cite{daniel2026lpwm}
& 20.7{\scriptsize$\pm$4.9}
& 37.8{\scriptsize$\pm$9.6}
& 73.5{\scriptsize$\pm$7.3}
& 65.4{\scriptsize$\pm$4.1}
& 36.2{\scriptsize$\pm$5.8} \\

& \textbf{ParticleSplat (Ours)}
& \textbf{54.1}{\scriptsize$\pm$4.7}
& \textbf{88.4}{\scriptsize$\pm$4.3}
& \textbf{90.6}{\scriptsize$\pm$6.9}
& \textbf{76.0}{\scriptsize$\pm$8.4}
& \textbf{83.8}{\scriptsize$\pm$4.6} \\

\midrule

\multirow{3}{*}{Multi-view}
& VAE-GS
& 34.2{\scriptsize$\pm$5.6}
& 79.3{\scriptsize$\pm$6.1}
& 62.8{\scriptsize$\pm$7.4}
& 57.1{\scriptsize$\pm$8.6}
& 72.9{\scriptsize$\pm$6.8} \\

& DLP~\cite{daniel2026lpwm}
& 41.7{\scriptsize$\pm$5.1}
& 63.4{\scriptsize$\pm$8.7}
& 83.6{\scriptsize$\pm$6.9}
& 66.8{\scriptsize$\pm$4.5}
& 48.5{\scriptsize$\pm$6.2} \\

& \textbf{ParticleSplat (Ours)}
& \textbf{82.6}{\scriptsize$\pm$4.3}
& \textbf{91.2}{\scriptsize$\pm$4.9}
& \textbf{94.1}{\scriptsize$\pm$6.2}
& \textbf{86.9}{\scriptsize$\pm$7.6}
& \textbf{91.4}{\scriptsize$\pm$4.2} \\
\bottomrule
\end{tabular*}

\vspace{0.5em}

\begin{tabular*}{\textwidth}{@{\extracolsep{\fill}} l l ccccccc @{}}
\toprule
\textbf{Setting} & \textbf{Method}/ \textit{Task}
& \begin{tabular}[c]{@{}c@{}}\textit{slide}\\\textit{block}\end{tabular}
& \begin{tabular}[c]{@{}c@{}}\textit{put in}\\\textit{drawer}\end{tabular}
& \begin{tabular}[c]{@{}c@{}}\textit{drag}\\\textit{stick}\end{tabular}
& \begin{tabular}[c]{@{}c@{}}\textit{push}\\\textit{buttons}\end{tabular}
& \begin{tabular}[c]{@{}c@{}}\textit{stack}\\\textit{blocks}\end{tabular}
& \multicolumn{1}{c}{\textbf{Overall}} \\
\midrule

\multirow{3}{*}{Single-view}
& VAE-GS
& 25.7{\scriptsize$\pm$10.9}
& 12.1{\scriptsize$\pm$3.0}
& 90.6{\scriptsize$\pm$7.8}
& 22.3{\scriptsize$\pm$11.4}
& 8.6{\scriptsize$\pm$3.0}
& 42.0 \\

& DLP~\cite{daniel2026lpwm}
& 54.6{\scriptsize$\pm$5.2}
& 30.8{\scriptsize$\pm$4.7}
& 93.1{\scriptsize$\pm$5.5}
& 36.7{\scriptsize$\pm$8.9}
& 11.4{\scriptsize$\pm$4.6}
& 46.6 \\

& \textbf{ParticleSplat (Ours)}
& \textbf{92.3}{\scriptsize$\pm$2.0}
& \textbf{78.4}{\scriptsize$\pm$5.8}
& \textbf{96.7}{\scriptsize$\pm$2.5}
& \textbf{57.2}{\scriptsize$\pm$7.1}
& \textbf{19.4}{\scriptsize$\pm$2.9}
& \textbf{73.0} \\

\midrule

\multirow{3}{*}{Multi-view}
& VAE-GS
& 47.9{\scriptsize$\pm$11.2}
& 31.4{\scriptsize$\pm$3.6}
& 92.1{\scriptsize$\pm$8.4}
& 39.8{\scriptsize$\pm$10.7}
& 9.7{\scriptsize$\pm$3.5}
& 52.7 \\

& DLP~\cite{daniel2026lpwm}
& 69.3{\scriptsize$\pm$5.6}
& 56.6{\scriptsize$\pm$5.1}
& 95.0{\scriptsize$\pm$5.0}
& 54.1{\scriptsize$\pm$9.3}
& 15.2{\scriptsize$\pm$4.2}
& 59.4 \\

& \textbf{ParticleSplat (Ours)}
& \textbf{94.2}{\scriptsize$\pm$2.4}
& \textbf{86.3}{\scriptsize$\pm$6.1}
& \textbf{96.1}{\scriptsize$\pm$2.9}
& \textbf{82.7}{\scriptsize$\pm$6.6}
& \textbf{24.8}{\scriptsize$\pm$3.2}
& \textbf{83.0} \\
\bottomrule
\end{tabular*}

\caption{\textbf{Ablation on representation design under single-view and multi-view settings.} We compare DLP, VAE-GS, and ParticleSplat on RLBench manipulation tasks under different view settings. Numbers denote success rate (\%) as mean $\pm$ standard deviation over three seeds (25 episodes per task per seed).}
\label{tab:ablation_repr}
\vspace{-2.5em}
\end{table*}

\subsubsection{A Recipe for Real-World Data Collection}
\label{subsec:apndx_recipe}

We propose a simple iPhone-based pipeline mimicking Structure-from-Motion (SfM~\cite{ullman1979sfm}) datasets, providing accurate point clouds via known intrinsics and extrinsics. Unlike RLBench's direct access, real-world RGB videos lack this information. We bridge this using ARKit\footnote{\url{https://developer.apple.com/augmented-reality/arkit/}}~\cite{arkit2024}, which tracks device pose via visual-inertial odometry in a gravity-aligned world frame. Intrinsics come from hardware specs. Table~\ref{tab:calibration_fields} details all ARKit outputs.

Our implementation\footnote{Tested on iPhone~15 and Apple~M2~Pro (requires iOS~11.0+, A9+ processor). Code will be released publicly.}, covering Xcode\footnote{\url{https://developer.apple.com/xcode/}}~\cite{Xcode} setup, for extraction of intrinsics and extrinsics, through video capture guidelines, is outlined in Fig.~\ref{fig:arkit_workflow}.

\begin{figure}[!ht]
    \centering
    \begin{lstlisting}[style=mypython, basicstyle=\ttfamily\tiny]
class ARSessionManager:
    calibration_stream : AsyncStream<CalibrationData>  # live intrinsics/extrinsics
    video_stream       : AsyncStream<URL>              # final MP4 path on stop
    calibration_frames : list[CalibrationData] = []    # per-written-frame record
    is_recording       : bool = False

def init():
    """Launch ARWorldTracking with gravity alignment and depth support"""
    session.run(ARWorldTrackingConfiguration())

def start_recording():
    """Initialize video writer; begin paired frame+calibration capture"""
    calibration_frames.clear()
    video_writer = AudioVisualAssetWriter()
    video_writer.start_session()
    attach(FRAME_UPDATED_EVENT, process_updates_callback)
    self.is_recording = True

def stop_recording() -> tuple[URL, list[CalibrationData]]:
    """Finalize MP4, emit path, return synchronized calibration frames"""
    self.is_recording = False
    video_writer.finish()  # async disk flush
    video_stream.yield(output_url)
    return output_url, calibration_frames

def process_updates_callback(frame: ARFrame):
    """ARKit delegate (~60Hz): extract RGB+calibration per frame"""
    rgb, calibration = process_frame(frame)
    calibration_stream.yield(calibration)  # live feed
    if self.is_recording:
        if video_writer.append(rgb, timestamp=frame.timestamp):
            calibration.frame_index = len(calibration_frames)
            calibration_frames.append(calibration)

def process_frame(frame: ARFrame) -> tuple[CGImage, CalibrationData]:
    """Extract synchronized RGB image and calibration data"""
    rgb = frame.captured_image
    cal = CalibrationData(
        intrinsics  = frame.camera.intrinsics,    # 3x3
        transform   = frame.camera.transform,     # 4x4 camera-to-world
        projection  = frame.camera.projection,    # 4x4
        euler       = frame.camera.euler_angles,
        timestamp   = frame.timestamp
    )
    return rgb, cal
    \end{lstlisting}
    \caption{\textbf{High-level ARKit camera manager pseudocode.} The delegate processes frames at $\sim$60Hz, recording calibration only for successfully-written video frames to ensure synchronization.}
    \label{fig:arkit_workflow}
\end{figure}

\begin{table*}[h]
    \centering
    \renewcommand{\arraystretch}{1.3}
    \begin{tabular}{p{0.25\textwidth} p{0.65\textwidth}}
        \hline
        \textbf{Field} & \textbf{Description} \\
        \hline
        Intrinsic Matrix & 3×3 affine encoding focal lengths and principal point \\
        Principal Point & Optical axis intersection with image plane (pixels) \\
        Focal Length & Hardware-calibrated (pixels) \\
        Transform Matrix & 4×4 camera-to-world pose (translation + rotation) \\
        Translation & Camera origin offset in world coordinates (m) \\
        Rotation Matrix & 3×3 orientation in world frame \\
        Euler Angles & Roll/pitch/yaw parameterization (rad) \\
        Projection Matrix & 4×4 world-to-image homogeneous mapping \\
        \hline
    \end{tabular}
    \caption{\textbf{ARKit calibration fields.} All data synchronized per video frame for accurate 3D reconstruction.}
    \label{tab:calibration_fields}
\end{table*}
 
\end{document}